\documentclass{article}

\usepackage{microtype}
\usepackage{graphicx}
\usepackage{booktabs} % for professional tables

\usepackage{hyperref}

\usepackage[accepted]{icml2021}

\icmltitlerunning{Learning Curves for Analysis of Deep Networks}

\usepackage{amsmath,amsfonts,bm}

\def\eqref#1{equation~\ref{#1}}
\def\1{\bm{1}}

\DeclareMathAlphabet{\mathsfit}{\encodingdefault}{\sfdefault}{m}{sl}
\SetMathAlphabet{\mathsfit}{bold}{\encodingdefault}{\sfdefault}{bx}{n}

\usepackage{times}
\usepackage{epsfig}
\usepackage{graphicx}
\usepackage{amsmath}
\usepackage{amssymb}
\usepackage{float}

\usepackage{url}
\usepackage{booktabs}
\usepackage{enumerate}
\usepackage{color}
\usepackage{jeffe} % Jeff's algorithm box.
\usepackage{subcaption}
\usepackage{multirow}
\usepackage{makecell}
\usepackage{caption}
\usepackage{footnote} % for footnote within tabular
\usepackage{xcolor}
\usepackage{xspace}
\usepackage{wrapfig}
\usepackage{hyperref}
\usepackage{cleveref}

\newcommand{\best}[1]{\underline{\textbf{#1}}}

\newcommand{\eg}{\emph{e.g.}\@\xspace}

\newcommand{\qbox}{\framebox[1em]{$\phantom{*}$}}
\newcommand{\hide}[1]{{\color[HTML]{D5D5D5}\colorbox[HTML]{F0F0F0}{\makebox[0.5em]{#1}}}}
\newcommand{\vect}[1]{\boldsymbol{#1}}

\makeatletter
\newcommand*{\storecounter}[2]{%
  \edef\@currentlabel{\the\value{#1}}% Store current counter value in \@currentlabel
  \label{#2}% Store label
}
\makeatother

\begin{document}

%%%%%%%%% TITLE
\twocolumn[
\icmltitle{Learning Curves for Analysis of Deep Networks}



\icmlsetsymbol{equal}{*}

\begin{icmlauthorlist}
\icmlauthor{Derek Hoiem}{uiuc}
\icmlauthor{Tanmay Gupta}{allen}
\icmlauthor{Zhizhong Li}{uiuc}
\icmlauthor{Michal M. Shlapentokh-Rothman}{uiuc}
%\icmlauthor{Iaesut Saoeu}{ed}
%\icmlauthor{Fiuea Rrrr}{to}
%\icmlauthor{Tateu H.~Yasehe}{ed,to,goo}
\end{icmlauthorlist}

%\icmlaffiliation{to}{Department of Computation, University of Torontoland, Torontoland, Canada}
\icmlaffiliation{uiuc}{University of Illinois at Urbana-Champaign}
\icmlaffiliation{allen}{PRIOR @ Allen Institute for AI}

\icmlcorrespondingauthor{Derek Hoiem}{dhoiem@illinois.edu}
%\icmlcorrespondingauthor{Eee Pppp}{ep@eden.co.uk}

\icmlkeywords{Evaluation, sample efficiency, deep learning}

\vskip 0.3in
]



\printAffiliationsAndNotice{}  % leave blank if no need to mention equal contribution

%\setlength{\tabcolsep}{5pt}

%\thispagestyle{empty}
%\pagestyle{plain}

\begin{abstract}
{\em Learning curves} model a classifier's test error as a function of the number of training samples. Prior works show that learning curves can be used to select model parameters and extrapolate performance. We investigate how to use learning curves to evaluate design choices, such as pretraining, architecture, and data augmentation. We propose a method to robustly estimate learning curves, abstract their parameters into error and data-reliance, and evaluate the effectiveness of different parameterizations. Our experiments exemplify use of learning curves for analysis and yield several interesting observations.
\end{abstract}

\section{Introduction}
\label{sec:intro}

%What gets measured gets optimized. We need better measures of learning ability to design better classifiers and predict the payoff of collecting more data. Currently, classifiers are evaluated and compared by measuring performance on one or more datasets according to a fixed train/test split. Ablation studies help evaluate the impact of design decisions. However, one of the most important characteristics of a classifier, how it performs with varying numbers of training samples, is rarely measured or modeled.  
The performance of a learning system depends strongly on the number of training samples, but standard evaluations use a fixed train/test split. While some works measure performance using subsets of training data, the lack of a systematic way to measure and report performance as a function of training size is a barrier to progress in machine learning research, particularly in areas like representation learning, data augmentation, and low-shot learning that specifically address the limited-data regime.  What gets measured gets optimized, so we need better measures of learning ability to design better classifiers for the spectrum of data availability.

\begin{figure*}[t!]
    \centering
    \includegraphics[width=\linewidth]{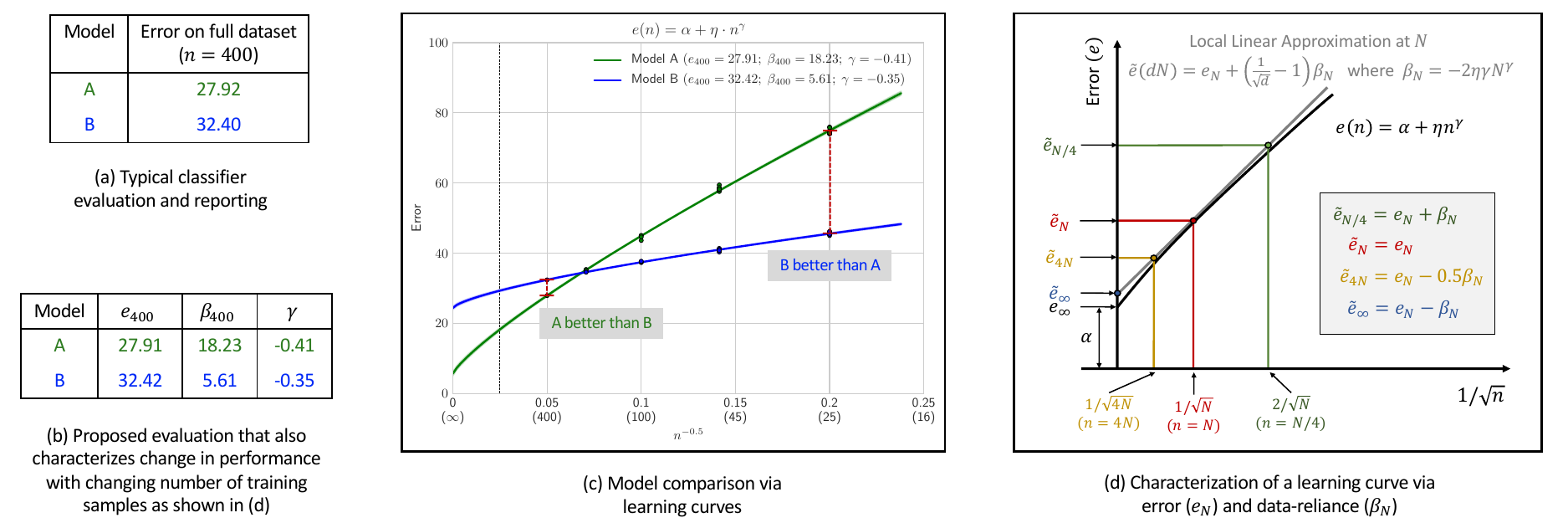}
    \vspace{-0.2in}
    \caption{
    \label{fig:teaser}
    \textbf{Evaluation with learning curves vs single-point comparison: } Comparing error of models trained on the full dataset, as shown in (a), is standard practice but provides an incomplete view of a classifier's performance.  We propose a methodology to estimate and visualize learning curves that model a classifier's performance with varying amounts of training data (b,c). We also propose a succinct summary of a model's curve in terms of error and data-reliance (b,d).
    %To better model how classifiers perform with varying amounts of training data, we propose a methodology to estimate and visualize learning curves (c), which can be succinctly summarized in terms of error $e_N$ and data-reliance $\beta_N$ (b, d).
    %and characterize learning curves (b,c)
    %how to estimate and visualizing learning curves that model a classifier's performance with varying amounts of training data (b,c). We also propose a succinct summary of a model's curve in terms of error and data-reliance (b,d).
    %\vspace{-0.1in}
    }
\end{figure*}

In this paper, we establish and demonstrate use of {\em learning curves}  to improve evaluation of classifiers (see Fig.~\ref{fig:teaser}).  
%refine the idea of {\em learning curves} that model error as a function of training set size. 
Learning curves, which model error as a function of training set size, were introduced nearly thirty years ago by \citet{cortes1993nips_learningcurves}) to accelerate model selection of deep networks, and recent works have demonstrated the predictability of performance improvements with more data~\citep{hestness_2017arxiv_deeplearningpredictable,johnson_2017_accuracyvstrainingsize,kaplan2020scaling,rosenfeld2020icml_generalizationError} or more network parameters~\citep{kaplan2020scaling,rosenfeld2020icml_generalizationError}.  But such studies aim to extrapolate rather than evaluate and have typically required large-scale experiments that are outside the computational budgets of many research groups. %, and their purpose is extrapolation rather than evaluation.  

Experimentally, we find that the extended power law  $e_{test}(n) = \alpha + \eta n^{\gamma}$ yields a well-fitting learning curve, where $e_{test}$ is test error and $n$ is the number of training samples (or ``training size''), but that the parameters \{$\alpha$, $\eta$, $\gamma$\} are individually unstable under measurement perturbations. To facilitate curve comparisons, we abstract the curve into two key parameters, $e_N$ and $\beta_N$, that have intuitive meanings and are more stable under measurement variance.  $e_N$ is the test error at $n=N$, and $\beta_N$ is a measure of data-reliance, how much a classifier's error will change if the training size changes.  Our experiments show that learning curves provide insights that cannot be obtained by single-point comparisons of performance. Our aim is to promote the use of learning curves as part of a standard learning system evaluation.

 %We develop a methodology and perform experiments that use learning curves to draw insights about deep convolutional network classifiers. Specifically, we can better understand how design decisions impact learning {\em potential} (measured as asymptotic error) and {\em efficiency} (measured as data sensitivity, the increase in error as number of training samples drops), rather than only how error changes for a particular training set size. This understanding can lead to better classifier designs and the ability to predict the benefit of collecting additional data.

%Quantity of training data is a primary factor in machine learning performance. Practitioners wonder what are the best design parameters for my current training data, and how much performance will improve with investment in more data collection and annotation? Currently, these questions are not easy to answer because most research evaluates on a single training set size or does not report in a way that supports extrapolation. 

Our \textbf{key contributions}:
\begin{itemize}
    \item Investigate how to best model, estimate, characterize, and display learning curves for  use in classifier analysis
    \item Exemplify use of learning curves with analysis of impact of error and data-reliance due to network architecture, optimization, depth, width, fine-tuning, data augmentation, and pretraining.
\end{itemize}

Table~\ref{tab:dl_quiz} shows validated and rejected popular beliefs that single-point comparisons often overlook.  In the following sections, we investigate how to model learning curves (Sec.~\ref{sec:model}), how to estimate them (Sec.~\ref{sec:estimation}), and what they can tell us about the impact of design decisions (Sec.~\ref{sec:experiments}), with discussion of limitations and future work in Sec.~\ref{sec:discussion}.
%and Appendix~\ref{app:discussion},\ref{app:user_guide}. %There is not yet extensive work on learning curves, so we weave related work discussions into the modeling and discussion sections. 

\begin{table*}[ht]
    \centering
    \resizebox{.99\textwidth}{!}{
    \begin{tabular}{lccc}
    \toprule
    \textbf{Popular beliefs}                                                                               & \begin{tabular}{c}Your\\ guess\end{tabular} & %\begin{tabular}{c}Our\\ guess\end{tabular} & 
    \begin{tabular}{c}Supp-\\orted?\end{tabular} & \begin{tabular}{c}Exp.\\figures\end{tabular} \\ \midrule
\emph{Pre-training} on similar domains nearly always helps compared to training from scratch.                                                       & \qbox %& \hide{T} 
                        & \hide{Y} & \ref{fig:imagenet_to_cifar},~\ref{fig:imagenet_to_places},~\ref{fig:exp_different_pretrained_models}\\
\emph{Pre-training}, even on similar domains, introduces bias that would harm performance with a large enough training set.                          & \qbox %& \hide{T} 
                      & \hide{U} & \ref{fig:exp_different_pretrained_models}\\
\emph{Self-/un-supervised training} performs better than supervised pre-training for small datasets.               
    & \qbox %& \hide{F} 
    & \hide{N} & \ref{fig:exp_different_pretrained_models}\\
\emph{Fine-tuning} the entire network (vs. just the classification layer) is only helpful if the training set is large.                             & \qbox %& \hide{T} 
        & \hide{N} & \ref{fig:imagenet_to_cifar},~\ref{fig:imagenet_to_places}\\
% \emph{Increasing network depth} harms performance for small training sets, due to an overly complex model.                                 & \qbox & \hide{T} & \hide{F} & \ref{fig:depth_pretr_ft},~\ref{fig:depth_pretr_linear}\\
% \emph{Increasing network depth} is more helpful for larger training sets than smaller ones.   & \qbox & \hide{T} & \hide{F} & \ref{fig:depth_pretr_ft},~\ref{fig:depth_pretr_linear}\\
\emph{Increasing network depth}, when fine-tuning, harms performance for small training sets, due to an overly complex model.                                 & \qbox %& \hide{T} 
            & \hide{N} & \ref{fig:depth_pretr_ft}\\
\emph{Increasing network depth}, when fine-tuning, is more helpful for larger training sets than smaller ones.   & \qbox %& \hide{T} 
            & \hide{N} & \ref{fig:depth_pretr_ft}\\
\emph{Increasing network depth}, if the backbone is frozen, is more helpful for smaller training sets than larger ones.   & \qbox %& \hide{T} 
        & \hide{N} & \ref{fig:depth_pretr_linear}\\
\emph{Increasing depth or width} improves more than ensembles of smaller networks with the same number of parameters. & \qbox %& \hide{T} 
        & \hide{Y} & \ref{fig:ensemble_vs_depth_vs_width}\\
%\emph{Increasing width} improves more than ensembles of smaller networks with same number of parameters.  & \qbox & \hide{T} & \hide{T} & \ref{fig:exp_ensemble},~\ref{fig:width_pretr_ft}\\
%\emph{Data augmentation} would not help with an infinite training set.                                     & \qbox & \hide{T} & \hide{F} & \ref{fig:aug_nopretr_ft},~\ref{fig:aug_pretr_ft}\\
\emph{Data augmentation} is roughly equivalent to using a $m$-times larger training set for some $m$.                                     & \qbox %& \hide{T} 
                & \hide{Y} & \ref{fig:aug}\\
%\emph{Data augmentation} is still important even if the backbone is frozen (i.e. not fine-tuning).           & \qbox & \hide{T} & \hide{F} & \ref{fig:aug}\\
%\emph{Pre-training} can perform worse than training from scratch when domains are different enough.           & \qbox & \hide{T} & \hide{?!} & fig\\ 
\bottomrule
    \end{tabular}
    }
    \caption{\textbf{Deep learning quiz!} We encourage our readers to judge each claim as T (true) or F (false) and then see if our experimental results support the claim (Yes/No/Unsure). In many cases, particularly regarding fine-tuning and network depth, the results surprised the authors. Our experiments show that learning curve analysis provides a systematic way to investigate these suppositions and others.
    } \vspace{-0.1in}
    \label{tab:dl_quiz}
\end{table*}

\section{Modeling Learning Curves}
\label{sec:model}

The learning curve measures test error $e_{test}$ as a function of the number of training samples $n$ for a given classification model and learning method. Previous empirical observations suggest a functional form $e_{test}(n) = \alpha + \eta n^\gamma$, with bias-variance trade-off and generalization theories typically indicating $\gamma=-0.5$.  We summarize what bias-variance trade-off and generalization theories (Sec.~\ref{sec:model_biasvariance}) and empirical studies (Sec.~\ref{sec:model_empirical}) can tell us about learning curves, and describe our proposed abstraction in Sec.~\ref{sec:model_ours}.

%Locally, e.g. in a range of $n=50$ to $n=400$, a simple model fits the curve: $e_{test}(n) = \alpha + \eta/\sqrt{n}$, where $\alpha$ is asymptotic error due to bias or non-unique input/output mappings, $\eta$ is the data reliance. Lower $\alpha$ means the classifier is better in the high-data regime, while lower $\eta$ means the classifier is better in the low-data regime (for similar $\alpha$). Theoretical support for this model comes from the bias-variance trade-off, as well as various generalization bounds. 

%Generalized power law model...
%The model is also supported by previous empirical studies, though with some caveats, as well as our own experiments.

\subsection{Bias-variance Trade-off and Generalization Theory}
\label{sec:model_biasvariance}
The bias-variance trade-off is an intuitive and theoretically sound way to think about generalization.  The ``bias'' is error due to inability of the classifier to encode the optimal decision function, and the ``variance'' is error due to %variations in predictions due to 
limited availability of training samples for parameter estimation.  This is called a trade-off because a classifier with more parameters tends to have less bias but higher variance. \citet{geman1992_neuralcomputation_biasvariance} decompose mean squared regression error into bias and variance and explore the implications for neural networks, leading to the conclusion that ``identifying the right preconditions is {\em the} substantial problem in neural modeling''. This conclusion foreshadows the importance of pretraining, though Geman et al. thought the preconditions must be built in rather than learned. \citet{domingos_2000_icml_biasvariance} extends the analysis to classification.
Theoretically, the mean squared error (MSE) can be modeled as $e_{test}^2(n) = bias^2 + noise^2 + var(n)$, where ``noise'' is irreducible error due to non-unique mapping from inputs to labels, and variance can be modeled as $var(n)=\sigma^2/n$ for $n$ training samples.

The $\eta n^\gamma$ term in $e_{test}(n)$ appears throughout machine learning generalization theory, usually with $\gamma=-0.5$.  For example, the bounds based on hypothesis VC-dimension~\citep{vapnik_1971_vcdim} and Rademacher Complexity~\citep{gnecco_2008_ams_rademacher} are both $O(c n^{-0.5})$ where $c$ depends on the complexity of the classification model. More recent work also follows this form, e.g.~\cite{neyshabur2018a,bartlettFT17,aroracompression18,bousquet02}. %We give some examples of bounds in Table~\ref{tab:bounds} without describing all of the parameters because the point is that the test error bounds vary with training size $n$ as a function of $n^{-0.5}$, for all approaches.     
One caveat is that the exponential term $\gamma$ can deviate from $-0.5$ if the classifier parameters depend directly on $n$.  For example, \cite{tsybakov2008_nonparametric} shows that setting the bandwidth of a kernel density estimator based on $n$ causes the dominant term to be bias, rather than variance, changing $\gamma$ in the bound.  In our experiments, all training and model parameters are fixed for each curve, except the learning schedule, but we note that the learning curve depends on optimization methods and hyperparameters as well as the classifier model.

\subsection{Empirical Studies}
\label{sec:model_empirical}
%While theory predicts that error should vary as a function of $n^{-0.5}$, theoretical predictions and empirical observations often have a loose relationship.  
Some recent empirical studies (e.g.~\citet{sun_iccv2017_unreasonable_effectiveness}) claim a log-linear relationship between error and training size, but this holds only when asymptotic error is zero.  \citet{hestness_2017arxiv_deeplearningpredictable} model error as $e_{test}(n) = \alpha + \eta n^\gamma$ but often find $\gamma$ much smaller in magnitude than $-0.5$ and suggest that poor fits indicate need for better hyperparameter tuning.   This empirically supports that data sensitivity depends both on the classification model and on the efficacy of the optimization algorithm and parameters. \citet{johnson_2017_accuracyvstrainingsize} also find a better fit with this extended power law model than by restricting $\gamma=-0.5$ or $\alpha=0$.  %We typically find an excellent learning curve fit with $-0.3 > \gamma > -0.7$. %using the Ranger optimizer~\citep{ranger}.

%we find that selecting the learning rate through validation on one training size and using the Ranger optimizer~\cite{ranger} leads to a consistently excellent fit with $\gamma=0.5$. %, to the extent that trying to estimate $\gamma$ leads to worse leave-one-out error in predicting error as a function of training size.

In the language domain, learning curves are used in a fascinating study by \citet{kaplan2020scaling}. For natural language transformers, they show that a power law relationship between logistic loss, model size, compute time, and dataset size is maintained if, and only if, each is increased in tandem. We draw some similar conclusions to their study, such as that increasing model size tends to improve performance {\em especially} for small training sets (which surprised us).  However, the studies are largely complementary, as we study convolutional nets in computer vision, classification error instead of logistic loss, and a broader range of design choices such as data augmentation, pretraining source, architecture, and optimization.  Also related, \citet{rosenfeld2020icml_generalizationError} model error as a function of both training size and number of model parameters with a five-parameter function that accounts for training size, model parameter size, and chance performance.  
A key difference in our work is that we focus on how to best draw insights about design choices from learning curves, rather than on extrapolation.  As such, we propose methods to estimate learning curves and their variance from a relatively small number of trained models.

\subsection{Proposed Characterization of Learning Curves for Evaluation}
\label{sec:model_ours}

Our experiments in Sec.~\ref{sec:validate} show that the learning curve model $e(n)=\alpha+\eta n^{\gamma}$ results in excellent leave-one-size-out RMS error and extrapolation. However, $\alpha$, $\eta$, and $\gamma$ cannot be meaningfully compared across curves because the parameters have high covariance with small data perturbations, and comparing $\eta$ values is not meaningful unless $\gamma$ is fixed and vice-versa. This would prevent tabular comparisons and makes it harder to draw quantitative conclusions.

%A classifier's performance can be characterized in terms of its error and data-reliance, or how quickly the error changes with training size $n$. With $e(n)=\alpha+\eta n^{\gamma}$, we find that $\gamma=-0.5$ provides a good local approximation but that fitting $\gamma$ significantly improves leave-one-size-out RMS error and extrapolation accuracy, as we detail in Sec.~\ref{sec:experiments}. However, $\alpha$, $\eta$, and $\gamma$ cannot be meaningfully compared across curves because the parameters have high covariance with small data perturbations, and comparing $\eta$ values is not meaningful unless $\gamma$ is fixed and vice-versa.

%One problem is that $\apha$, $\eta$, and $\gamma$ co-vary wildly with small perturbations of data, such that different combinations could produce similar curves near the range of observed training sizes.
%A dataset, network architecture, and learning algorithm (loss function, optimizer, and hyper-parameter selection strategy) defines a learning curve $e(n)=\alpha+\eta n^{\gamma}$ parameterized by $\alpha$ (asymptotic error), $\eta$ (inverse learning efficiency), and $\gamma$ (a parameter that determines the curvature of the curve). One possibility for evaluating the impact of changes in dataset, architecture, or learning algorithm is to directly compare the three parameters. However, $\alpha$ (error with infinite training data) is sensitive to choice of gamma and error measurements used to fit the curve, and comparing $\eta$s is only meaningful when $\alpha$s and $\gamma$s are similar across settings. 
To overcome this problem, we propose to report error and sensitivity to training size in a way that can be derived from various learning curve models and is insensitive to data perturbations.  The curve is characterized by error $e_N =\alpha+\eta N^{\gamma}$ and data-reliance $\beta_N$, and we typically choose $N$ as the full dataset size. Noting that most learning curves are locally well approximated by a model linear in $n^{-0.5}$, we compute data-reliance as $\beta_N =N^{-0.5}\left.\frac{\partial{e}}{\partial{n^{-0.5}}}\right|_{n=N}=-2\eta\gamma N^\gamma$.  When the error is plotted against $n^{-0.5}$, $\beta_{N}$ is the slope at $N$ scaled by $N^{-0.5}$, with the scaling chosen to make the practical implications of $\beta_N$ more intuitive. 
%Typically, the quantities of interest are error at full dataset size and the change in error if the current dataset size was increased or reduced by a factor (\eg~$4\times$, $2\times$, $\frac{1}{2}\times$, $\frac{1}{4}\times$). Therefore, given the full dataset size $n_f$, we recommend reporting ``error@$n_f$'': $e_{n_f}=\alpha+\eta n_f^{\gamma}$ and ``data-reliance'': $\beta_{n_f} = n_f^{-0.5}\left.\frac{\partial{e}}{\partial{n^{-0.5}}}\right|_{n=n_f}$. 
 This yields a simple predictor for error when changing training size by a factor of $d$:
%Linear approximation of the curve around $N$ on this plot results in a simple formula for computing change in error due to a $d\times$ change in dataset size:
\begin{equation}
    \widetilde{e}(d\cdot N) = e_N + \left(\frac{1}{\sqrt{d}}-1\right)\beta_N.
    \label{eq:linearized_e_estimate}
\end{equation}
This is a first order Taylor expansion of $e(n)$ around $n=N$ with respect to $n^{-0.5}$.
By this linearized estimate, asymptotic error is $e_N-\beta_N$, a 4-fold increase in data (\eg~$400\to 1600$) reduces error by $0.5\beta_N$, and using only one quarter of the dataset (\eg~$400\to 100$) increases the error by $\beta_N$. For two models with similar $e_N$, the one with a larger $\beta_N$ would outperform with more data but underperform with less. %However, when comparing curves with different $n_f$s (e.g. across datasets), we recommend using $e_n$ and $\beta_n$ at the same $n$. 
$(e_N,\beta_N,\gamma)$ is a complete re-parameterization of the extended power law, with $\gamma+0.5$ indicating the curvature in $n^{-0.5}$ scale. See Fig.~\ref{fig:teaser}d for illustration.

\section{Estimating Learning Curves}
\label{sec:estimation}

We now describe the method for estimating the learning curve from error measurements with confidence bounds on the estimate. %:et $e_{ij}$ denote the random variable corresponding to the test error when the model is trained on the $j^{\text{th}}$ fold.
%With subsampled folds of size $n_i$ (either per class or in total) out of $F_i\leq\lfloor\frac{N}{n_i}\rfloor$ total folds we use, let $e_{ij}$ denote the random variable corresponding to the test error when the model is trained on the $j^{\text{th}}$ fold. 
Let $e_{ij}$ denote the random variable corresponding to test error when the model is trained on the $j^{\text{th}}$ fold of $n_i$ samples (either per class or in total). 
We assume $\{e_{ij}\}_{j=1}^{F_i}$ are i.i.d according to $\mathcal{N}(\mu_i,\sigma_i^2)$. We want to estimate learning curve parameters $\alpha$ (asymptotic error), $\eta$, and $\gamma$, such that $e_{ij}=\alpha+\eta n_i^{\gamma}+\epsilon_{ij}$ where $\epsilon_{ij} \sim \mathcal{N}(0,\sigma_i^2)$ and $\mu_{ij}=\mathbb{E}[e_{ij}]=\mu_i$. Sections~\ref{sec:lsq_given_gamma} and~\ref{sec:curve_mean_var} describe how to estimate mean and variance of $\alpha$ and $\eta$ for a given $\gamma$, and Sec.~\ref{sec:est_gamma} describes our approach for estimating $\gamma$. 

\subsection{Weighted Least Squares Formulation}\label{sec:lsq_given_gamma}

We estimate learning curve parameters $\{\alpha,\eta\}$ by optimizing a weighted least squares objective:
\begin{equation}\label{eq:curve_estimation_objective}
    \mathcal{G}(\gamma) = \min_{\alpha,\eta}\sum_{i=1}^S\sum_{j=1}^{F_i} w_{ij}\left(e_{ij}-\alpha-\eta n^{\gamma}\right)^2
\end{equation}
where $w_{ij}=1/(F_i\sigma_i^2)$.  $F_i$ is the number of models trained with data size $n_i$ and is used to normalize the weight so that the total weight for observations from each training size does not depend on $F_i$. The factor of $\sigma_i^2$ accounts for the variance of $\epsilon_{ij}$.  Assuming constant $\sigma_i^2$ and removing the $F_i$ factor would yield unweighted least squares.

%Different assumptions about error variance could be easily substituted, for example $\sigma_i^2=\sigma^2/n_i$ or $\sigma_i^2$ is constant. 

%. For example, variance inversely proportional to dataset size \ie \;$\sigma_i^2=\sigma^2/n_i$ is equivalent to setting $w_{ij}=n_i/(F_i\sigma^2)$, and constant variance irrespective of dataset size \ie \; ${\sigma_i^2=\sigma}$ corresponds to $w_{ij}=1/(F_i\sigma^2)$, when still normalizing by $F_i$/

The variance of the estimate of $\sigma_i^2$ from $F_i$ samples is $2\sigma_i^4/F_i$, which can lead to over- or under-weighting data for particular $i$ if $F_i$ is small. Recall that each sample $e_{ij}$ requires training an entire model, so $F_i$ is always small in our experiments.  We would expect the variance to have the form ${\sigma}_i^2 = \sigma_0^2 + \hat\sigma^2/n_i$, where $\sigma_0^2$ is the variance due to random initialization and optimization and $\hat\sigma^2/n_i$ is the variance due to randomness in selecting $n_i$ samples. We validated this variance model by averaging over the variance estimates for many different network models on the CIFAR-100~\citep{cifar100} dataset.  We use $\sigma_0^2=0.02$ in all experiments and then use a least squares fit to estimate a single $\hat\sigma^2$ parameter from all samples $\mathbf{e}$ in a given learning curve. % as a least squares fit and also upper-bounds  $w_{ij}<=5$ even if two models happen to have the same error.  
This attention to $w_{ij}$ may seem fussy, but without such care we find that the learning curve often fails to account sufficiently for all the data in some cases.

\subsection{Solving for Learning Curve Mean and Variance}\label{sec:curve_mean_var}

Concatenating errors across dataset sizes (indexed by $i$) and folds results in an error vector $\vect{e}$ of dimension $D=\sum_{i=1}^SF_i$. For each $d\in\{1,\cdots,D\}$, $\vect{e}[d]$ is an observation of error at dataset size $n_{i_d}$ that follows $\mathcal{N}(\mu_{i_d},\sigma_{i_d}^2)$ with $i_d$ mapping $d$ to the corresponding $i$. %Fobe the total number of data points each consisting of $\{e_d,n_d,w_d,\mu_d,\sigma_d^2\}_{d=1}^D$, 

The weighted least squares problem can be formulated as solving a system of linear equations denoted by $W^{1/2}\vect{e}=W^{1/2}A\vect{\theta}$,
%\begin{equation}
%    W^{1/2}\vect{e}=W^{1/2}A\vect{\theta}
%\end{equation} 
where $W\in\mathbb{R}^{D\times D}$ is a diagonal matrix of weights ${W_{dd}=w_d}$, %$\vect{e}\in \mathbb{R}^D$ is a column vector of measured errors with $\vect{e}[d]\sim \mathcal{N}(\mu_{d(i)},\sigma_{d(i)}^2)$, 
$A\in\mathbb{R}^{D\times 2}$ is a matrix with ${A[d,:]=\left[1\;\;n_d^{\gamma}\right]}$, and $\vect{\theta}=\left[\alpha\;\;\eta\right]^T$ are the parameters of the learning curve, treating $\gamma$ as fixed for now.  The estimator for the learning curve is then given by $\hat{\vect{\theta}} = (W^{1/2}A)^+W^{1/2}\vect{e} = M\vect{e}$,
%\begin{equation}
%    \hat{\vect{\theta}} = (W^{1/2}A)^+W^{1/2}\vect{e} = M\vect{e}
%\end{equation} 
where $M\in\mathbb{R}^{2\times D}$ and $^+$ is pseudo-inverse operator. 

%We compute a mean curve using
%\begin{equation}
%    \overline{\vect{\theta}} = \mathbb{E}[\hat{\vect{\theta}}] = M\mathbb{E}[\vect{e}] = M\vect{\mu}
%\end{equation}
%where $\vect{\mu}\in \mathbb{R}^{D}$ with $\vect{\mu}[d]=\hat \mu_{i_d}$ computed by empirical estimate as $\sum_{j=1}^{F_i}e_{ij}/F_i$. 

The covariance of the estimator is given by 
%\begin{equation}
    $\Sigma_{\hat{\vect{\theta}}} = M\Sigma_{\vect{e}}M^T$, 
%\end{equation}
where $\Sigma_{\hat{\vect{\theta}}} \in \mathbb{R}^{2\times 2}$ and $\Sigma_{\vect{e}}\in\mathbb{R}^{D\times D}$ is the diagonal covariance of $\vect{e}$ with  $\Sigma_{\vect{e}}[d,d]=\sigma_{i_{d}}^2$. 
We compute our empirical estimate of $\sigma_i^2$ as described in Sec.~\ref{sec:lsq_given_gamma}.
%use $\hat{\sigma}_i^2$ described in  
%as an empirical estimate of $\sigma_i^2$. 

Since the estimated curve is given by $\hat{e}(n)=\left[1\;\;n^{\gamma}\right]\vect{\hat{\theta}}$, %the mean curve can be computed as
%\begin{equation}
%$    \overline{e}(n) = 
%    \begin{bmatrix}
%    1 & n^{\gamma}
%    \end{bmatrix}\overline{\vect{\theta}} = 
%    \overline{\alpha} + \overline{\eta}n^{\gamma}$.
%\end{equation}
the $95\%$ bounds at any $n$ can be computed as ${\hat{e}(n) \pm 1.96\times\hat{\sigma}(n)}$ with
\begin{align}
    \hat{\sigma}^2(n) = 
    \begin{bmatrix}
    1 & n^{\gamma}
    \end{bmatrix}
    \Sigma_{\hat{\vect{\theta}}}
    \begin{bmatrix}
    1 \\
    n^{\gamma}
    \end{bmatrix}
\end{align}
%\begin{align}
%    \hat{\sigma}(n) &= \text{Var}(\hat{\alpha}) + \frac{\text{Var}(\hat{\eta})}{n} + 2\frac{\text{Cov}(\hat{\alpha},\hat{\eta})}{\sqrt{n}} \\
%    &= \Sigma_{\hat{\theta}}[1,1] + \frac{\Sigma_{\hat{\theta}}[2,2]}{n} + 2\frac{\Sigma_{\hat{\theta}}[1,2]}{\sqrt{n}}
%\end{align}
%where $\hat\alpha$ and $\hat\eta$ are the empirical estimates of $\alpha$ and $\eta$.
%\input{figs/effect_of_wts}
%\input{figs/effect_of_ns}
For a given $\gamma$, these confidence bounds reflect the variance in the estimated curve due to variance in error measurements.

%the sample variance, the confidence that a single model trained with $n_i$ data points will have an error within confidence bounds of the mean learning curve, assuming there exists a curve with fixed $\gamma$ that fits the true mean.  
%variance in the curve due to variance in error measurements. %sample variance, i.e. reflecting the confidence that a single model trained with $n_i$ data points will have an error within confidence bounds of the mean learning curve, assuming there exists a curve with fixed $\gamma$ that fits the true mean.  

\subsection{Estimating $\gamma$}\label{sec:est_gamma}
We search for $\gamma$ that minimizes the weighted least squares objective with an L1-prior that slightly encourages values close to $0.5$. Specifically, we solve
\begin{equation}
    \min_{\gamma \in \left(-1,0\right)} \mathcal{G}(\gamma) + \lambda |\gamma+0.5|
\end{equation}
by searching over $\gamma\in\{-0.99,..., -0.01\}$ with $\lambda=5$ for our experiments.

\section{Experiments}
\label{sec:experiments}

\begin{figure*}[h!]
\centering
\begin{subfigure}[c]{0.62\textwidth}
\resizebox{\textwidth}{!}{%
\begin{tabular}{lcccccccccc}
\toprule
& & & \multicolumn{7}{c}{RMSE}   \\
\cmidrule{4-10} % \cmidrule{12-18}
\textbf{Params} & \textbf{Weights} & $R^2$  & \textbf{25} & \textbf{50} & \textbf{100} & \textbf{200} & \textbf{400} & \textbf{avg} & \textbf{p-value}  \\
\midrule
\multirow{3}{*}{$\alpha,\eta, \gamma$} 
& $\frac{1}{\sigma_i^2F_i}$ & 0.998 & 2.40 & 0.86 & \best{0.54} & 0.57 & \best{0.85} & \best{1.04} & -  \\
& $\frac{1}{\sigma_i^2}$ & 0.999 & \best{2.38} & 0.83 & 0.69 & 0.54 & 1.08 & 1.10 & 0.06  \\
& $1$ & 0.998 & 2.66 & 0.86 & 0.79 & \best{0.50} & 1.26 & 1.21 & 0.008   \\
\cmidrule{1-10}
$\alpha, \eta$
& $\frac{1}{\sigma_i^2F_i}$ & 0.988 & 3.41 & 1.09 & 0.69 & 0.72 & 1.21 & 1.42 & \textless{}0.001   \\
\cmidrule{1-10}
$\alpha, \eta, \delta$
& $\frac{1}{\sigma_i^2F_i}$ & 0.999 & 2.89 & \best{0.74} & 0.68 & 0.56 & 0.94 & 1.16 & 0.05  \\
\cmidrule{1-10}
$\alpha, \eta, \delta, \gamma$
& $\frac{1}{\sigma_i^2F_i}$ & 0.999 & 3.46 & \best{0.74} & 0.70 & 0.59 & 1.00 & 1.30 & 0.02  \\
\bottomrule
\end{tabular}
}
\end{subfigure} 
\begin{subfigure}[c]{0.32\textwidth}
    \includegraphics[width=\textwidth]{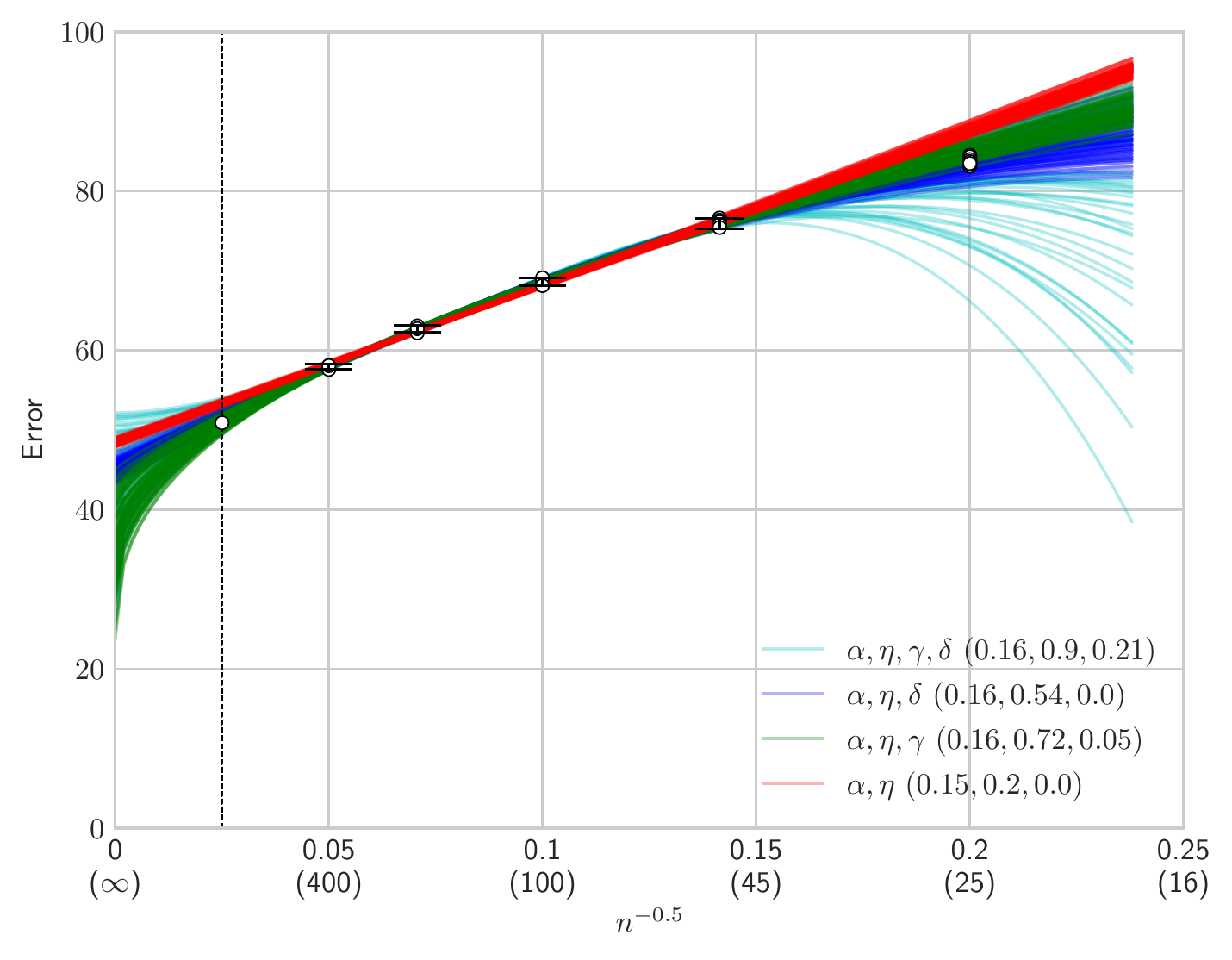}
\end{subfigure}
\caption{\label{fig:validity}
\textbf{Learning curve model and weights validation.} See Sec.~\ref{sec:validate} for explanation.  
}
\end{figure*}

We validate our choice of learning curve model and estimation method in 
Sec.~\ref{sec:validate} and use the learning curves to explore impact of design decisions on error and data-reliance in Sec.~\ref{sec:exp_curves}.

%Sec.~\ref{sec:implementation}, apply learning curves to gain insights about error and data-reliance in Sec.~\ref{sec:exp_curves}, and validate our choice of learning curve parameterization and fitting weights used in the least squares objective in Sec.~\ref{sec:validate}.  

\textbf{Setup: } Each learning curve is fit to test errors measured after training the classifier on various subsets of the training data. When less than the full training set is used, we train multiple classifiers using different partitions of data (e.g. four classifiers on four quarters of the data). We avoid use of benchmark test sets, since the curves are used for analysis and ablation. Instead, training data are split 80/20, with 80\% of data used for training and validation and remaining 20\% for testing.
%Since the curves are used for analysis and ablation, the test error is computed on a held out validation set (not the benchmark's test set). 
For each curve, all models are trained using an initial learning rate that is selected using one subset of training data, and the learning rate schedule is set for each $n$ based on validation. Unless otherwise noted, models are pretrained on ImageNet. Other hyperparameters are fixed for all experiments. Unless otherwise noted, we use the Ranger optimizer~\citep{ranger}, 
%version 20.4.11, 
which combines Rectified Adam~\citep{liu2019variance}, Look Ahead~\citep{Zhang2019LookaheadOK}, and Gradient Centralization~\citep{Yong2020GradientCA}, as preliminary experiments showed its effectiveness.  For ``linear'', we train only the final classification layer with the other weights frozen to initialized values.  All weights are trained when ``fine-tuning''. %Except in Fig.~\ref{fig:additional_datasets}, 
Tests are on Cifar100~\cite{cifar100}, Cifar10, Places365~\citep{zhou2017places}, or Caltech-101~\cite{caltech101_FeiFeiFergusPeronaPAMI}.  See supplemental materials (Appendix~\ref{supp:implementation}) for more implementation details.

\subsection{Evaluation of Learning Curves Model and Fitting}\label{sec:validate}

We validate our learning curve model using leave-one-size-out prediction error, e.g. predicting empirical mean performance with 400 samples per class based on observing error from models trained on 25, 50, 100, and 200 samples.   %We consider various choices of weighting schemes ($w$'s in Eq.~\ref{eq:curve_estimation_objective}) and estimating different parameters in a general form of the learning curve given by $e(n) = \alpha+\eta n^\gamma + \delta n^{2\gamma}$. Setting $\delta=0$ yields the model used in our experiments. 

\noindent\textbf{Weighting Schemes.} In the Fig.~\ref{fig:validity} table, learning curve prediction error is averaged for 16 Cifar100 classifiers with varying design decisions.  We compare 
three weighting schemes ($w$'s in Eq.~\ref{eq:curve_estimation_objective}): $w_{ij}=1$ is unweighted; $w_{ij}=1/\sigma_i^2$ is weighted by estimated size-dependent standard deviation; $w_{ij}=1/(F_i\sigma_i^2)$ makes the total weight for a given dataset size invariant to the number of folds. On average our proposed weighting performs best with high significance compared to unweighted. The p-value is paired t-test of difference of means calculated across all dataset sizes. 

\noindent\textbf{Model Choice.} We consider other parameterizations that are special cases of $e(n)=\alpha + \eta n^\gamma + \delta n^{2\gamma}$. Setting $\delta=0$ (top row of table) yields the model described in Sec.~\ref{sec:model_ours} and used in our experiments. The table in Fig.~\ref{fig:validity} shows that our model outperforms the others, in most cases with high significance, and achieves a very good fit with $R^2$ of 0.998.    

\noindent\textbf{Model Stability.} Each data point requires training a classifier, so we want to verify whether the curves can be estimated from few points.  We test stability and sample requirements by repeatedly fitting curves to four resampled data points for a model (Resnet-18, no pretraining, fine-tuned, tested on Places365).  Based on estimates of mean and standard deviation, one point each at $n=\{50, 100, 200, 400\}$ is sampled and used to fit a curve, repeated 100 times.  Parentheses in legend show standard deviation of estimates of $e_N$, $\beta_N$, and $\gamma$. Our preferred model extrapolates best to $n=1600$ and $n=25$ (plotted as white circles) while retaining stable estimates of of $e_N$ and $\beta_N$, but predicted asymptotic error $\alpha$ varies widely.  Supplemental material (Appendix~\ref{app:big_table}) shows similar estimates of $e_N$ and $\beta_N$ by fixing $\gamma=-0.5$ and fitting only $\alpha$ and $\eta$ on the three largest sizes (typically $n=\{100,200,400\}$), indicating that a lightweight approach of training a few models can yield similar conclusions.

\subsection{Learning Curve Comparisons}
\label{sec:exp_curves}

We explore a broad range of design decisions, aiming to exemplify the use of learning curves and demonstrate that such analysis leads interesting observations. Most of these design decisions warrant more complete investigation in separate papers to draw more general conclusions. %, and we show how learning curves can play an important role in those investigations. 

\begin{figure}[h!]
    \centering
    \begin{subfigure}[b]{0.4\textwidth}
        \includegraphics[width=\textwidth]{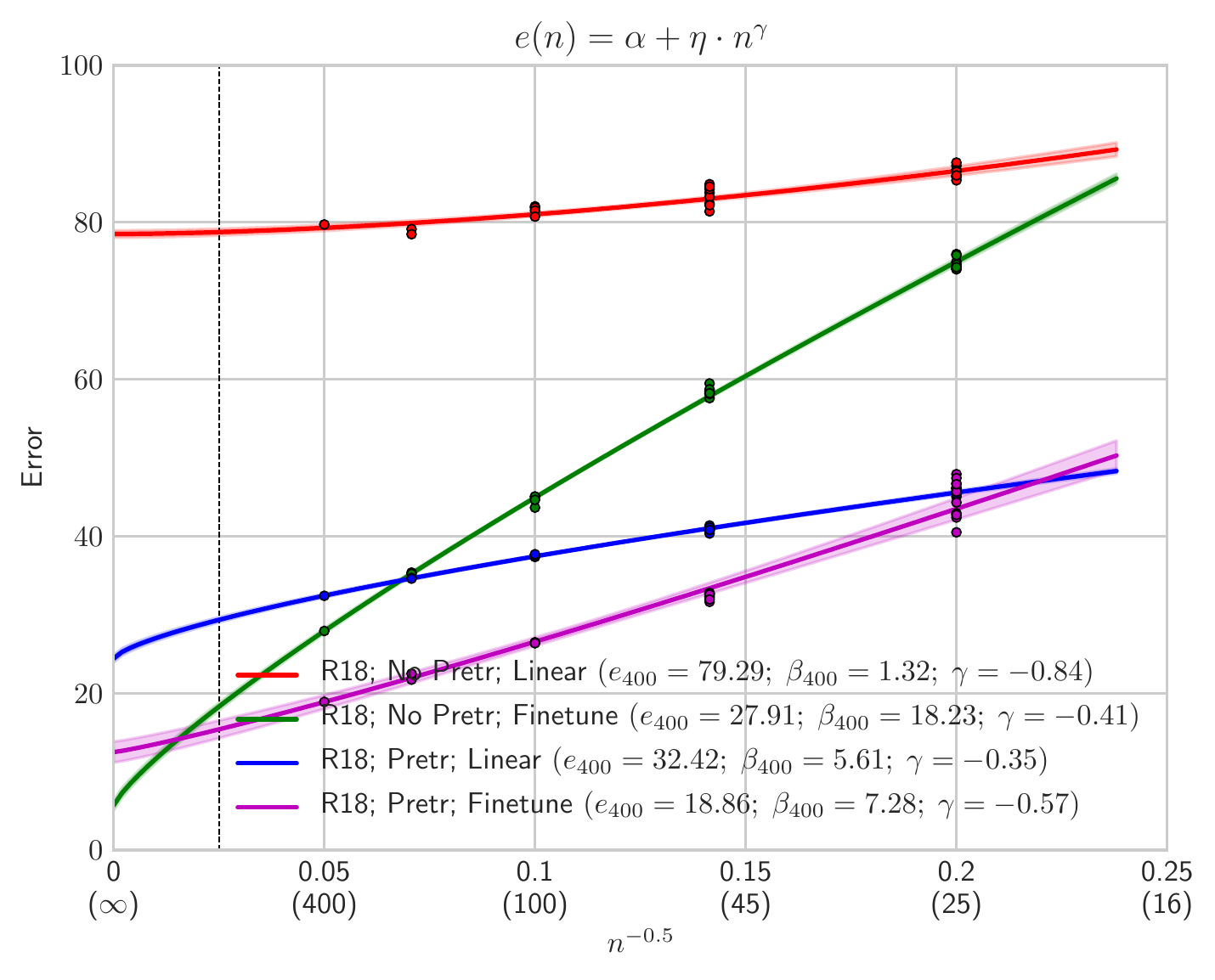}
        \caption{Transfer: ImageNet to Cifar100}
        \label{fig:imagenet_to_cifar}
    \end{subfigure} \\
    \begin{subfigure}[b]{0.4\textwidth}
        \includegraphics[width=\textwidth]{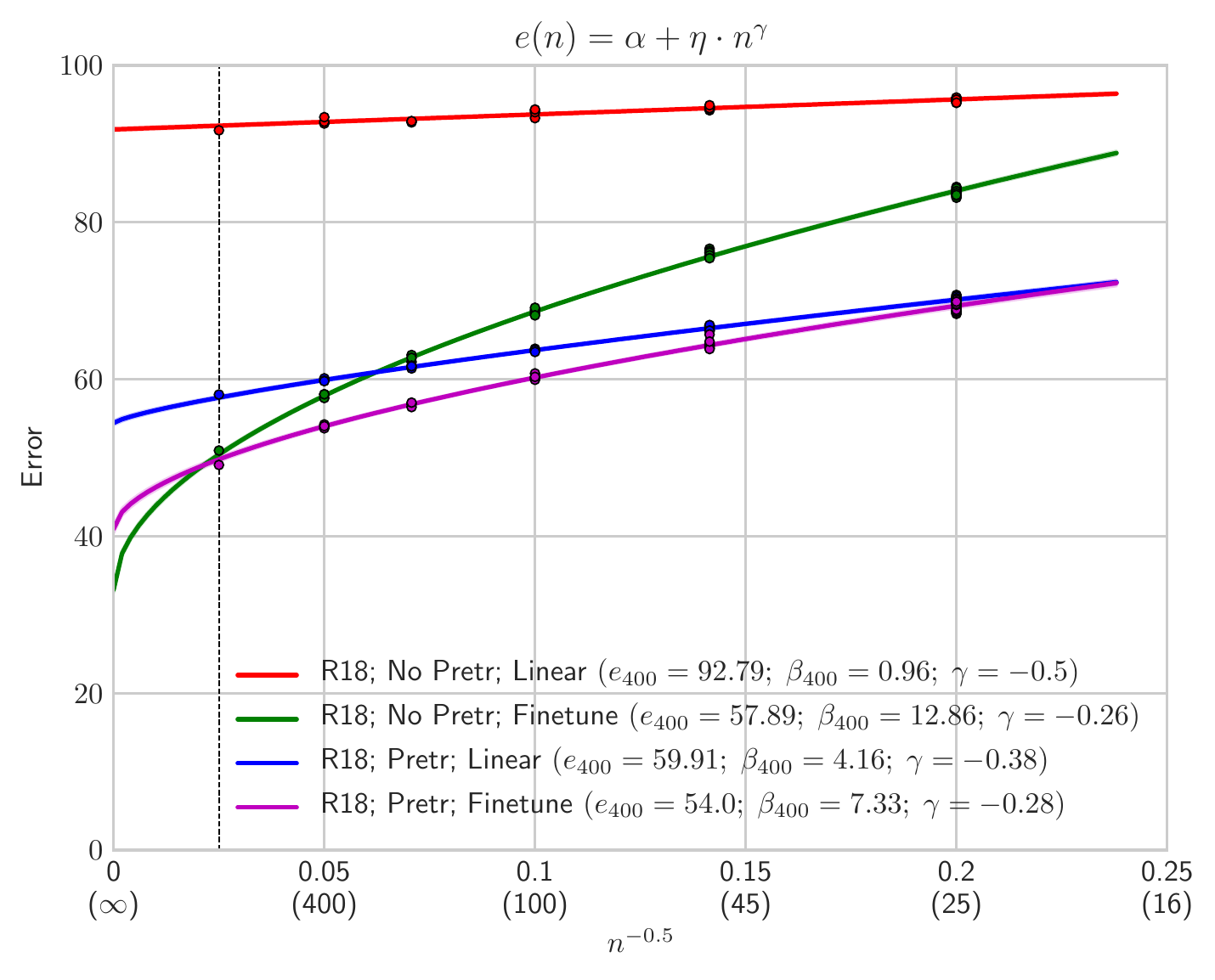}
        \caption{Transfer: ImageNet to Places365}
        \label{fig:imagenet_to_places}
    \end{subfigure}
\caption{\label{fig:initialization}\textbf{Pretraining and fine-tuning} with ResNet-18.}
\vspace{-0.1in}
\end{figure}

%We hope these experiments will inspire more complete explorations in each topic, as well as routine use for evaluation of new architectures, optimization methods, data augmentation methods, pre-trained representations, and so on.
\textbf{Figures: } We plot the fitted learning curves and confidence bounds, with observed test errors as circles. The legend displays \textbf{$\gamma$}, \textbf{error $e_N$}, and \textbf{data reliance $\beta_N$} with $N=400$ for Cifar100 and Places365 and $N=4000$ for Cifar10.  The x-axis is in scale $n^{-0.5}$ ($n$ in parentheses is the number of samples per class), but $\gamma$ is fit for each curve.  
%except Fig.~\ref{fig:additional_datasets} where $n$ is the total number of samples. 
A vertical bar indicates $n=1600$, which we consider the limit of accurate extrapolation from curves fit to $n\leq 400$ samples. All points are used for fitting, except in Fig.~\ref{fig:imagenet_to_places} $n=1600$ is held out to test extrapolation. {\em Best viewed in color.}

\begin{figure}[h!]
        \vspace{-0.15in}
        \includegraphics[width=\columnwidth]{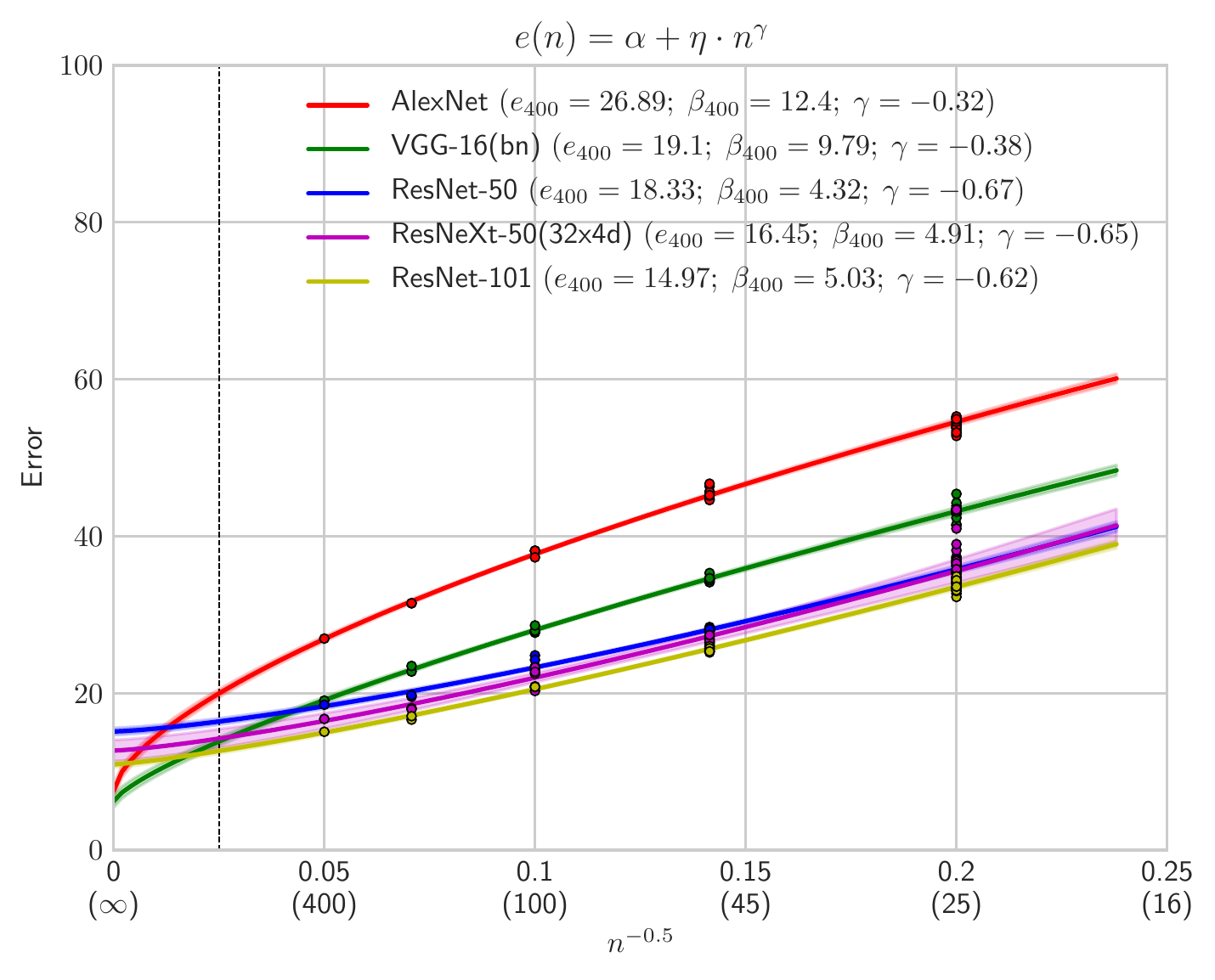}
        \vspace{-0.3in}
        \caption{\label{fig:arch}\textbf{ Architecture} (Cifar100 w/ finetuning)}
        \vspace{-0.25in}
\end{figure}

\textbf{Network architecture: }
%\input{figs/effect_of_arch}
%Do the newer architectures dominate, or are there some regimes in which the older architectures still outperform?
Advances in CNN architectures have reduced number of parameters while also reducing error over the range of training sizes. On Cifar100, AlexNet has 61M parameters; VGG-16, 138M; ResNet-50, 26M; ResNeXt-50, 25M; and ResNet-101, 45M. The landmark architecture papers do not examine effect of training data size, so it is interesting to see in Fig.~\ref{fig:arch} that each major advance through ResNet reduces both data reliance and $e_{400}$. ResNeXt appears to slightly reduce $e_{400}$ without change to data reliance.

% \begin{figure*}[h!]
%     \centering
%     \begin{subfigure}[b]{0.48\textwidth}
%         \includegraphics[width=\textwidth]{imgs/opt_no_pretr_ft.pdf}
%         \caption{without Pretrain}
%         \label{fig:opt_no_pretrain}
%     \end{subfigure}
%     \begin{subfigure}[b]{0.48\textwidth}
%         \includegraphics[width=\textwidth]{imgs/opt_pretr_ft.pdf}
%         \caption{with ImageNet Pretrain}
%         \label{fig:opt_pretrain}
%     \end{subfigure}
% \caption{\label{fig:optimization}\textbf{Optimization} on Cifar10 with ResNet-18.}
% \end{figure*}

\begin{figure*}[h!]
    \centering
    \begin{subfigure}[b]{0.4\textwidth}
        \includegraphics[width=\textwidth]{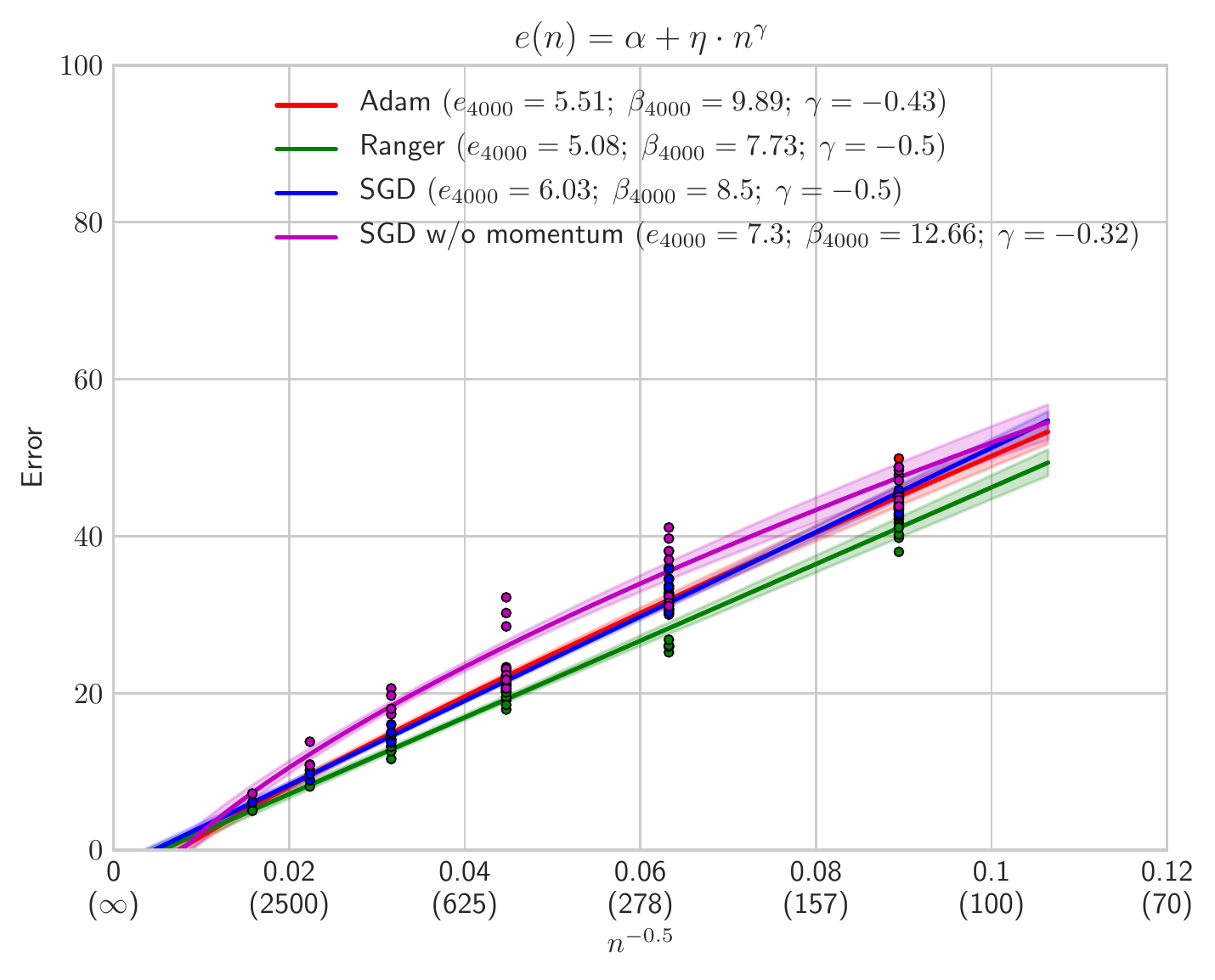}
        %\caption{without Pretrain}
        \label{fig:opt_no_pretrain}
    \end{subfigure} 
    \begin{subfigure}[b]{0.4\textwidth}
        \includegraphics[width=\textwidth]{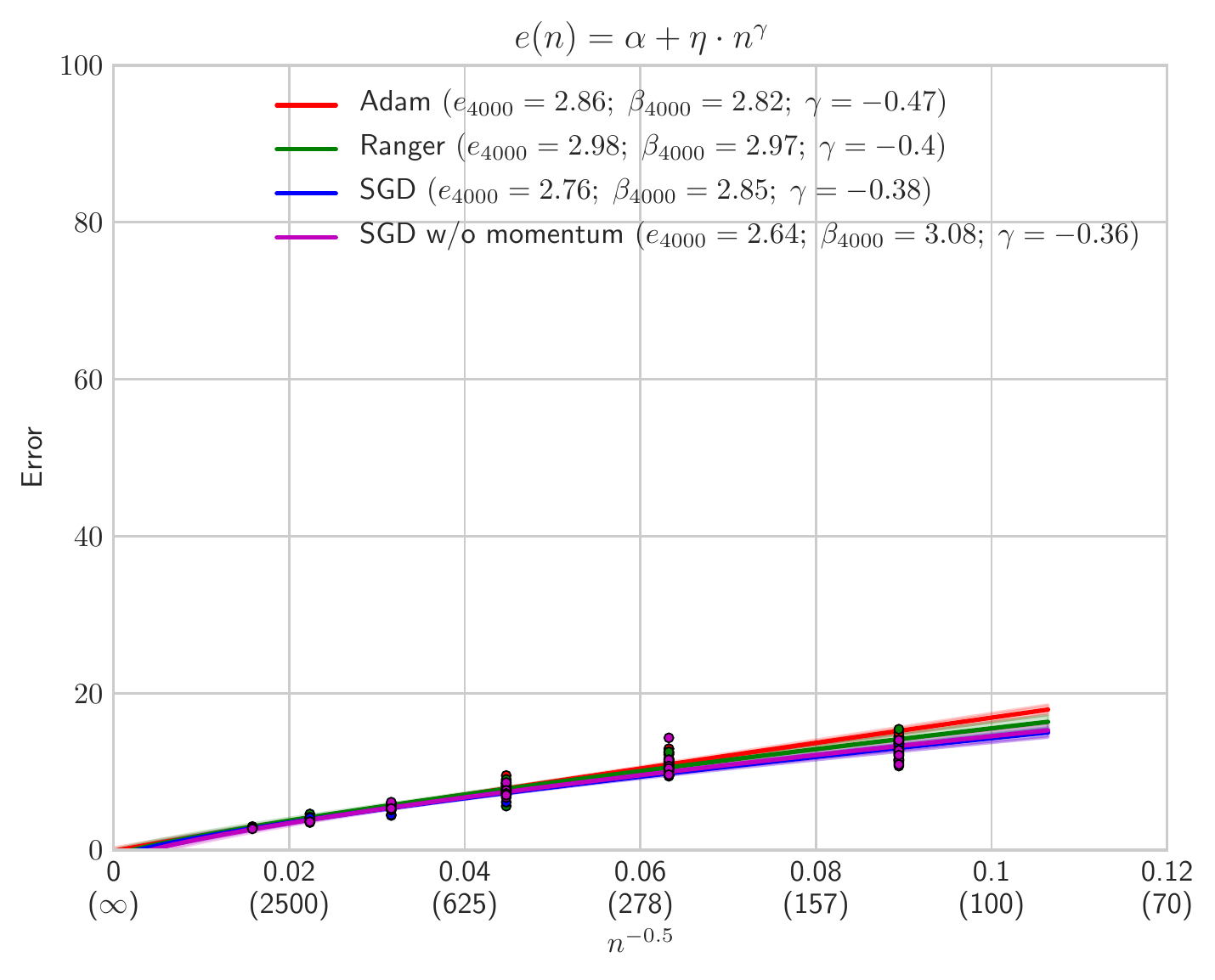}
        %\caption{with ImageNet Pretrain}
        \label{fig:opt_pretrain}
    \end{subfigure} \vspace{-0.3in}
\caption{\label{fig:optimization}\textbf{Optimization} on Cifar10 with ResNet-18. Without pretraining on left; with ImageNet pretraining on right.}
\vspace{-0.15in}
\end{figure*}
\textbf{Optimization method: }
In Fig.~\ref{fig:optimization}, we show results on Cifar10 when training ResNet-18 using four different optimization methods: Ranger~\citep{ranger}, Adam~\citep{kingma_iclr2015_adam}, stochastic gradient descent (SGD) w/ momentum, and SGD w/o momentum. %The training set is split to use up to 4000 samples per class for training/validation, and 1000 per class for testing. 
%Similarly to our experiments with Cifar100, we use 80\% of the standard training set for training and validation (4000 examples per class) and the remaining 20\% for testing.  
With pretraining, all methods perform similarly, but when training from scratch Ranger outperforms with lower $e_{4000}$ and $\beta_{4000}$.  SGD without momentum performs the worst and is least consistent across folds. Optimization papers routinely show error as a function of training iterations, but not the relationship to training size, so it is interesting to see empirically how the optimizer matters most with small data sizes and/or no pretraining, likely because better optimizers reduce variance of parameter estimation.

\textbf{Pretraining and fine-tuning: }  %However, we cannot draw conclusions about bias because extrapolated asymptotic error is not reliable, and the full story is complicated.
%but pretraining has a large effect on data-reliance $\beta_{400}$. Thus, pretraining does not bias the classifier, but improves its sample efficiency.
%This supports the finding of \cite{he2019rethinking} that pre-traininer is a ``warm start'' that can be overcome with enough training, where He et al. find a randomly initialized network can approach a pre-trained model with a long learning schedule.
Some prior works examine the effectiveness of pretraining. \citet{kornblith_2019cvpr_pretraining} show that  fine-tuned pretrained models outperform randomly initialized models across many architectures and datasets, but the gap is often small and narrows as training size grows.  For object detection, \citet{he2019rethinking} find that, with long learning schedules, randomly initialized networks approach the performance of networks pretrained for classification, even on smaller training sizes, and  \citet{zoph_2020arxiv_selftraining} show that pretraining can sometimes harm performance when strong data augmentation is used.   %All agree that pretraining is useful for cases of very limited data and for reducing training time. 

Compared to prior works, our use of learning curve models enables  extrapolation beyond available data and numerical comparison of data reliance. In Fig.~\ref{fig:initialization} we see that, without fine-tuning (``linear''), pretraining leads to a huge improvement in $e_{400}$ for all training sizes.  When fine-tuning, the pretraining greatly reduces data-reliance $\beta_{400}$ and also reduces $e_{400}$, providing strong advantages with smaller training sizes that may disappear with enough data.

%\begin{figure}[h]
%\includegraphics[width=\linewidth]{imgs/effect_of_pretr/imagenet_to_cifar.png}
%\caption{\textbf{Effects of pretraining.} Learning curves for Resnet-18 trained on Cifar-100 with and without pretraining while either finetuning all layers or the last classification layer.}
%\end{figure}

% \begin{figure}[h!]
%     \centering
%     \begin{subfigure}[b]{0.48\linewidth}
%         \includegraphics[width=\textwidth]{imgs/pretr_in2cifar.pdf}
%         \caption{Transfer: Imagenet to Cifar100}
%         \label{fig:imagenet_to_cifar}
%     \end{subfigure}
%     \begin{subfigure}[b]{0.48\linewidth}
%         \includegraphics[width=\textwidth]{imgs/pretr_in2places.pdf}
%         \caption{Transfer: Imagenet to Places365}
%         \label{fig:imagenet_to_places}
%     \end{subfigure}
% \caption{\label{fig:initialization}\textbf{Pretraining and fine-tuning} with ResNet-18.}
% \end{figure}
\begin{figure*}[h!]
    \centering
    %\vspace{-0.15in}
    \begin{subfigure}[b]{0.4\textwidth}
        \includegraphics[width=\textwidth]{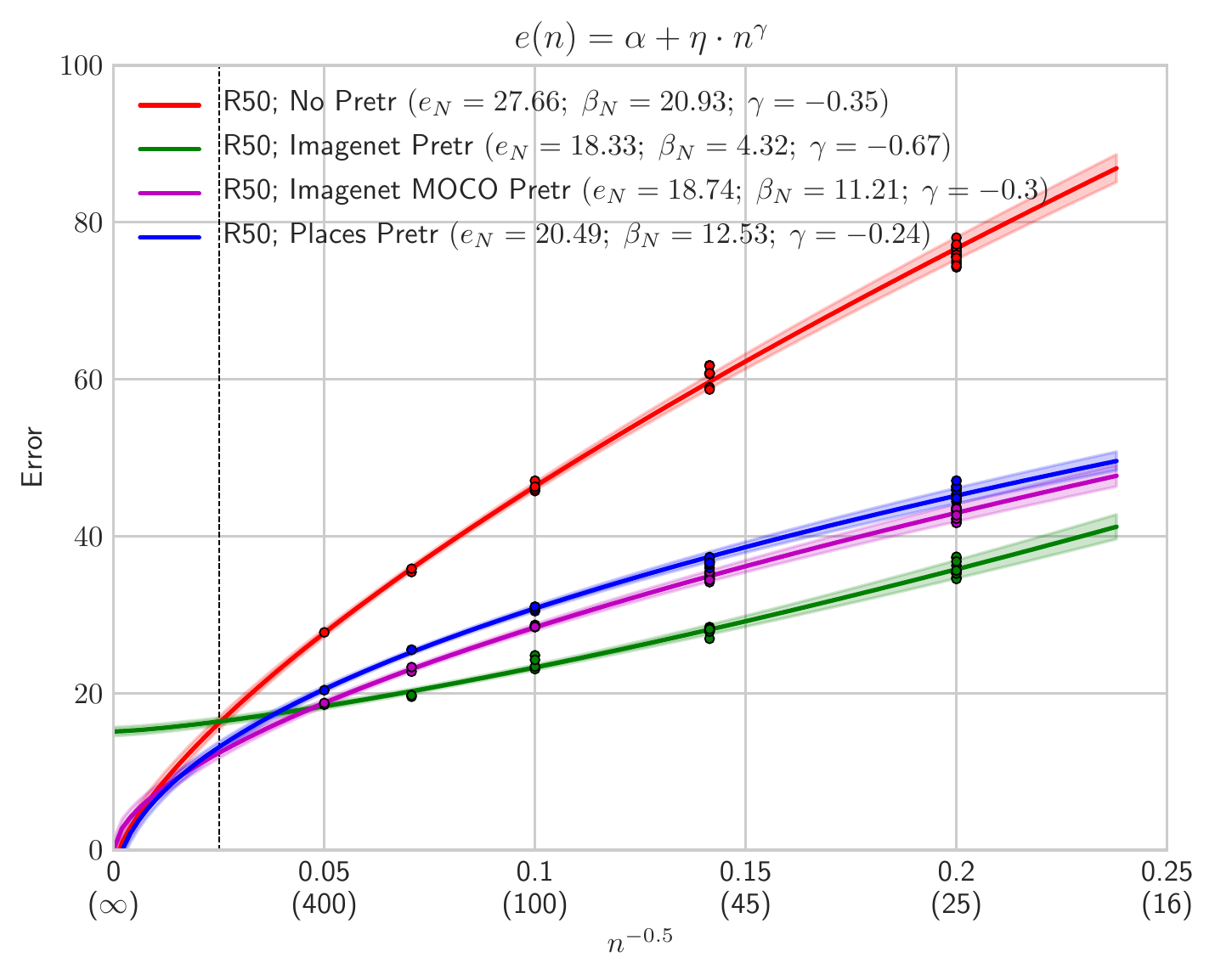}
        %\caption{Test on Cifar100}
        %\label{fig:sources_cifar}
    \end{subfigure} 
    \begin{subfigure}[b]{0.4\textwidth}
        \includegraphics[width=\textwidth]{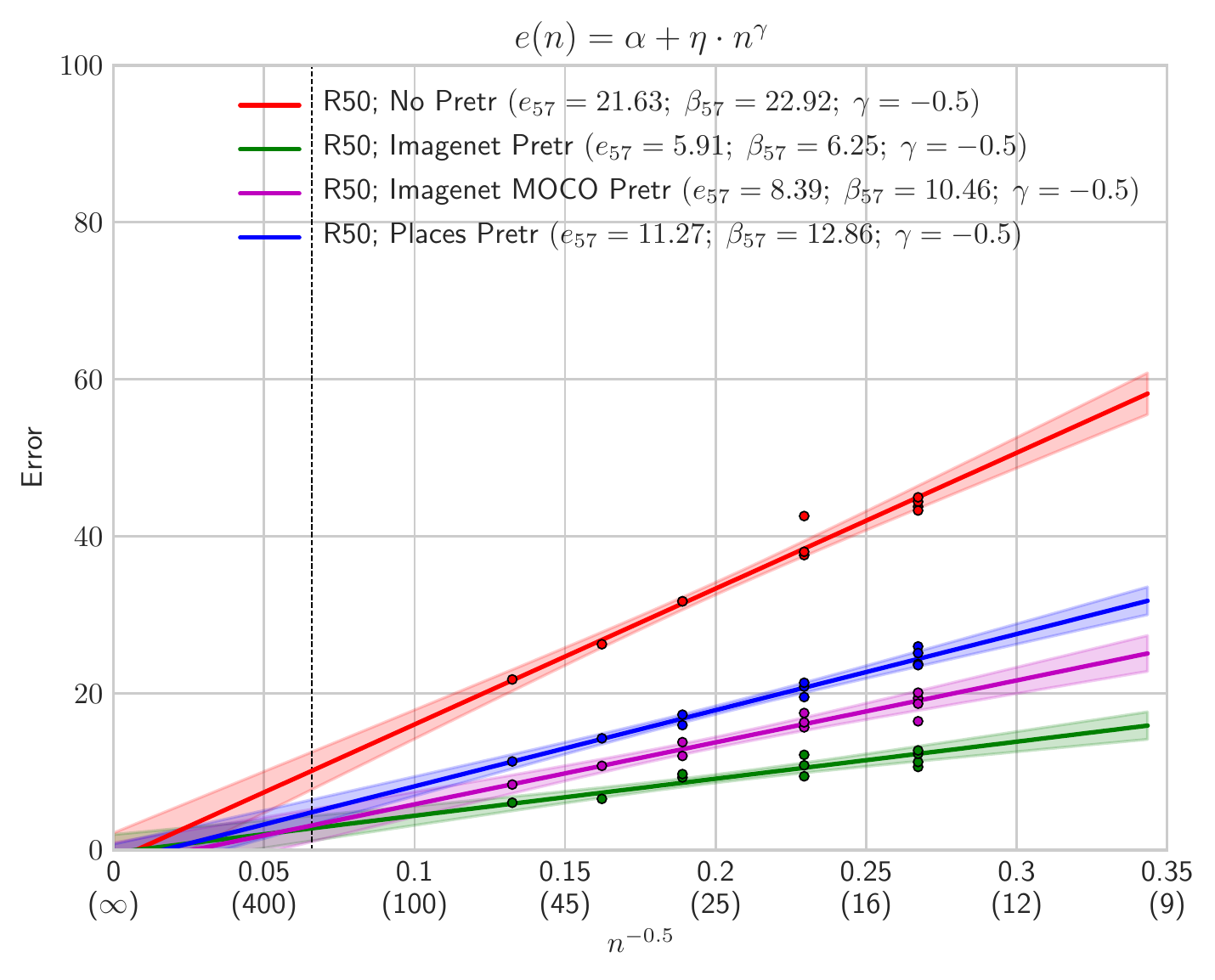}
        %\caption{Test on Caltech-101}
        %\label{fig:sources_caltech}
    \end{subfigure} \vspace{-0.1in}
\caption{\label{fig:exp_different_pretrained_models}\textbf{Pretraining sources}: test on Cifar100 (left) and Caltech-101 (right).} \vspace{-0.15in}
\end{figure*}

\textbf{Pretraining data sources: } In Fig.~\ref{fig:exp_different_pretrained_models}, we test on Cifar100 and Caltech-101 to compare different pretrained models: randomly initialized, supervised on ImageNet or Places365~\cite{zhou2017places}, and self-supervised on ImageNet (MOCO by~\citet{he2020moco}). The strongest impact is on data-reliance, which leads to consistent orderings of models across training sizes for both datasets. Supervised ImageNet pretraining has lowest $e_{N}$ and $\beta_{N}$, then self-supervised MOCO, then supervised Places365, with random initialization trailing far behind all pretrained models.
%All initializations have similar extrapolated error at $n=1600$, but different data-reliance.  
%Self-supervised MOCO leads to lower $e_{N}$ and $\beta_{N}$ compared to supervised Places365 pretraining, and  supervised pretraining on ImageNet has the lowest $e_{N}$ and $\beta_{N}$. 
The original papers excluded either analysis of training size (MOCO) or fine-tuning (Places365), so ours is the first analysis on image benchmarks to compare finetuned models from different pretraining sources as a function of training size. Newly proposed methods for representation learning would benefit from further learning curve analysis.

%We suspect that the $\gamma=-0.67$ and higher extrapolated asymptotic error may be due to measurement noise and suboptimal hyperparameter selection. 

\begin{figure*}[ht]
    \captionsetup[subfigure]{font=footnotesize,labelfont=scriptsize}
    \centering
    % depth: ft
    \begin{subfigure}[t]{0.31\textwidth}
        \includegraphics[width=\textwidth]{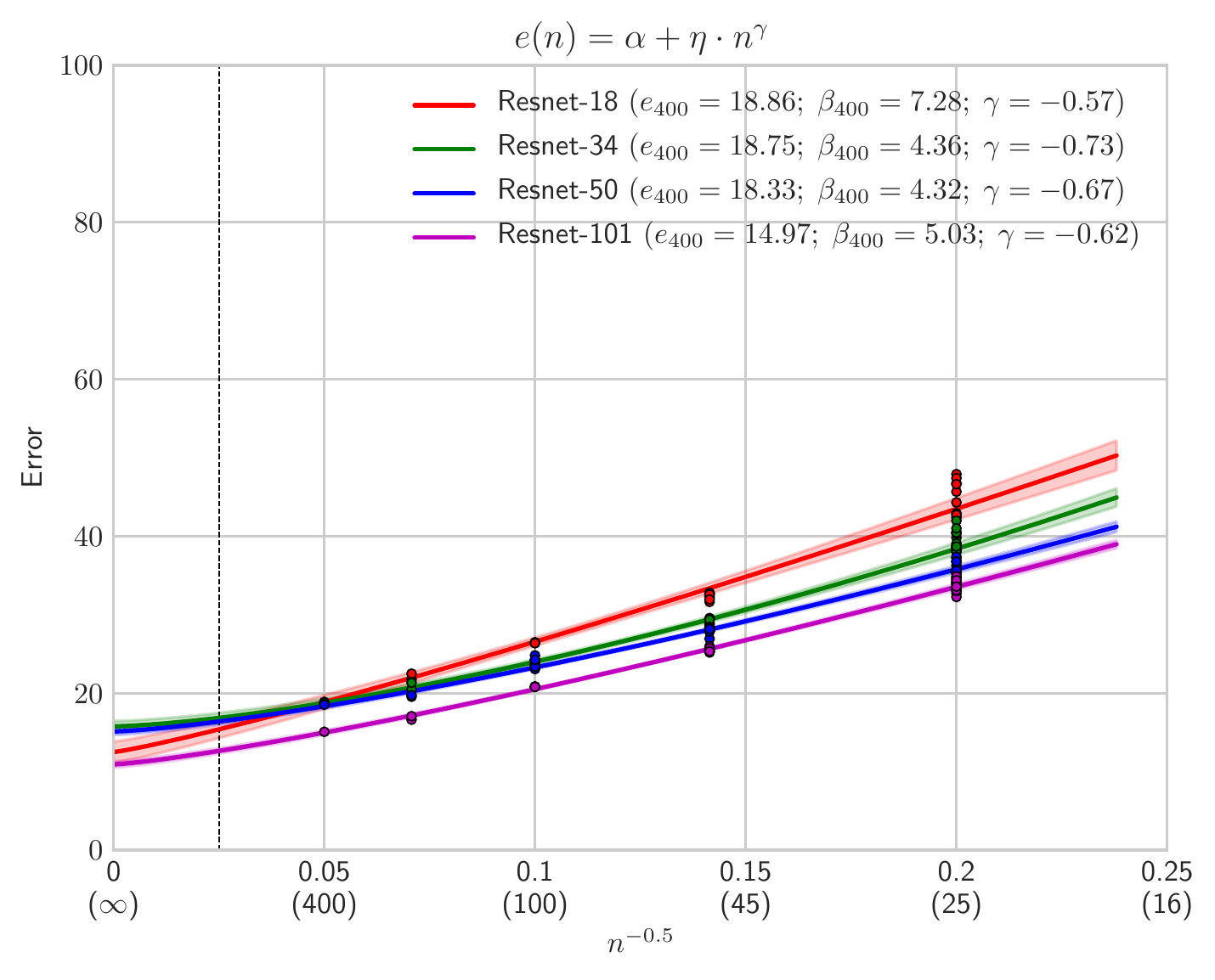}
        \caption{Depth: finetune}
        \label{fig:depth_pretr_ft}
    \end{subfigure}
    % width: ft
    \begin{subfigure}[t]{0.31\textwidth}
        \includegraphics[width=\textwidth]{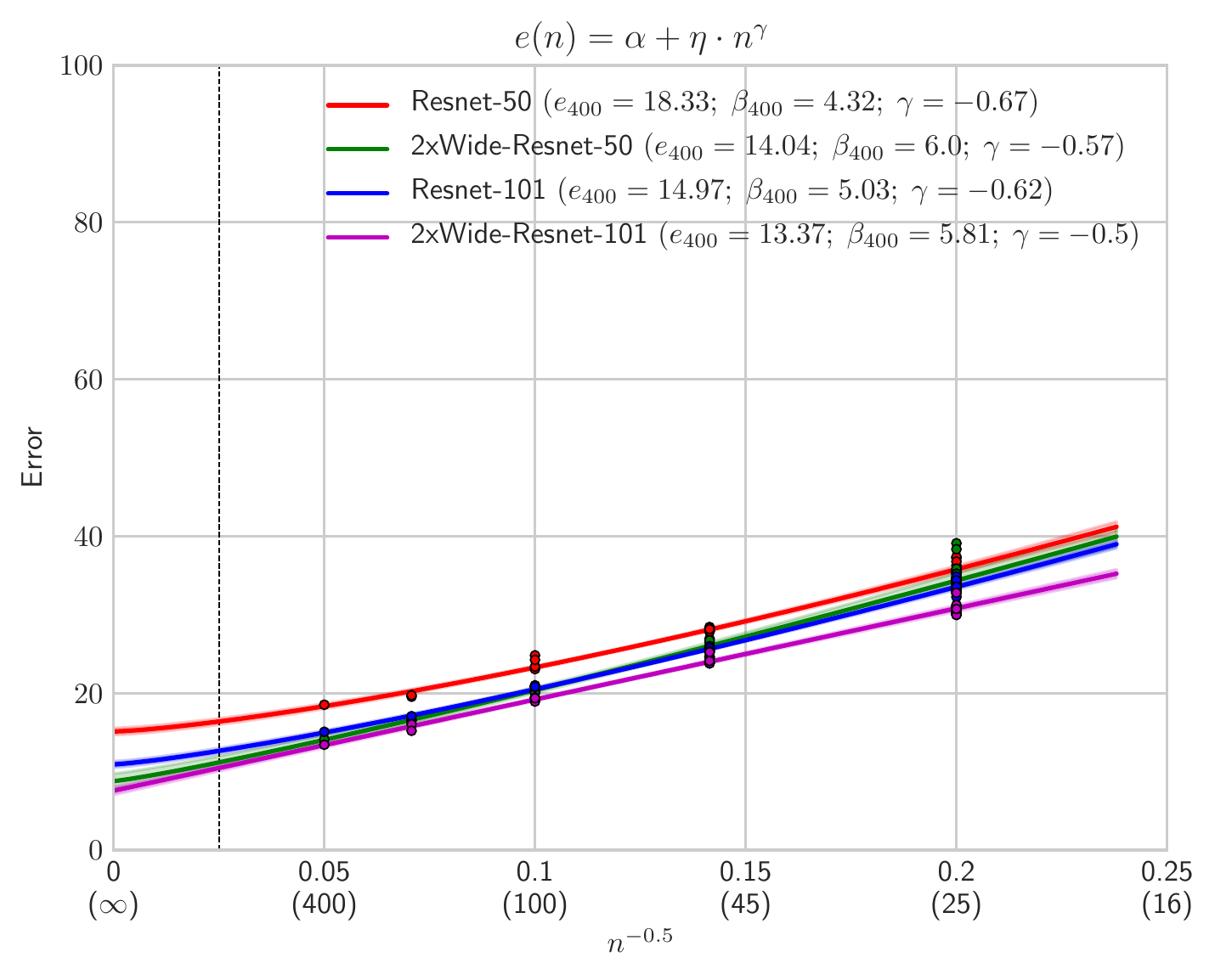}
        \caption{Width: finetune}
        \label{fig:width_pretr_ft}
    \end{subfigure}
    % ensemble: ft
    \begin{subfigure}[t]{0.31\textwidth}
        \includegraphics[width=\textwidth]{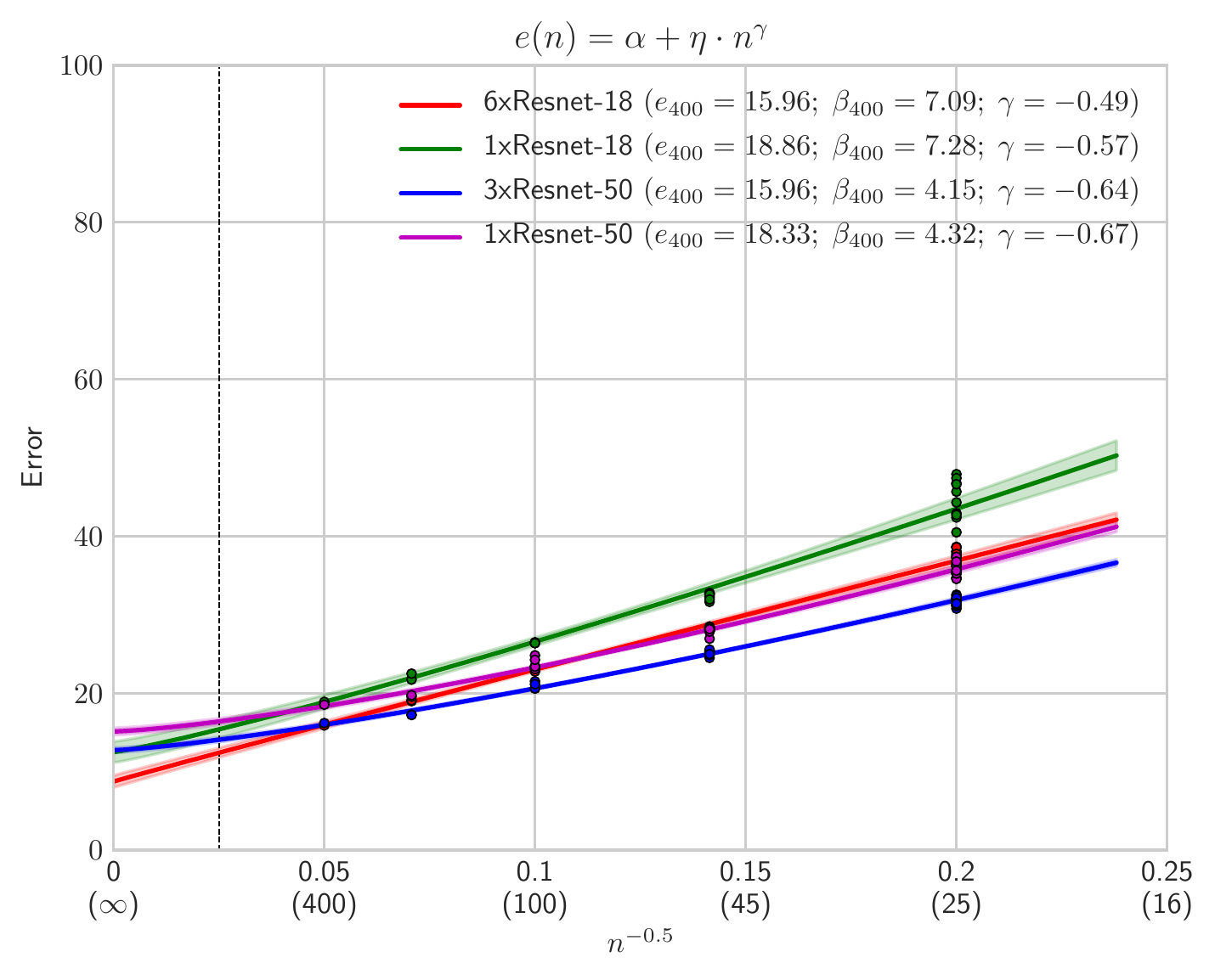}
        \caption{Ensemble: finetune}
        \label{fig:exp_ensemble}
    \end{subfigure}
    \\
    % depth: lin
    \begin{subfigure}[t]{0.31\textwidth}
        \includegraphics[width=\textwidth]{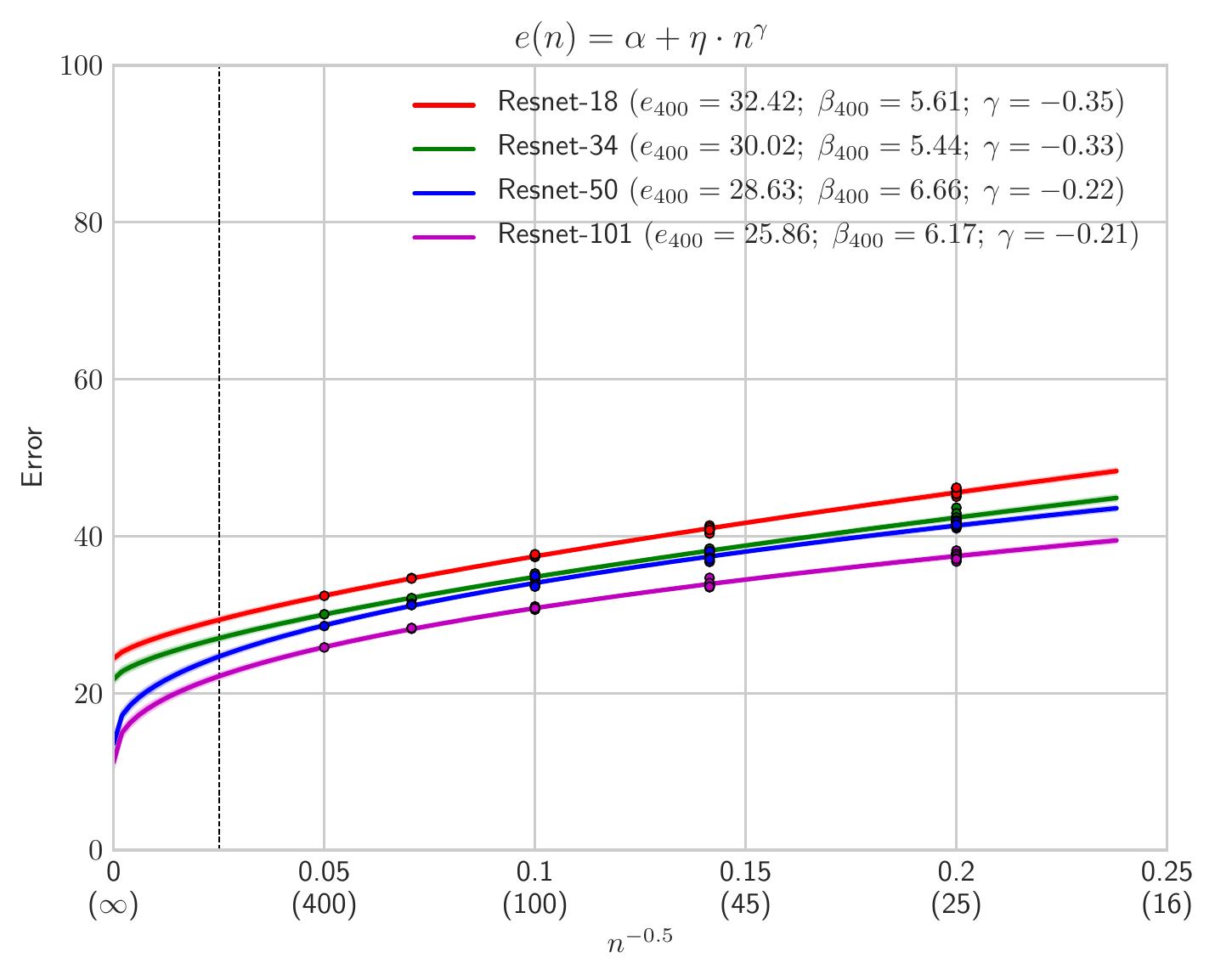}
        \caption{Depth: linear}
        \label{fig:depth_pretr_linear}
    \end{subfigure}
    % width: lin
    \begin{subfigure}[t]{0.31\textwidth}
        \includegraphics[width=\textwidth]{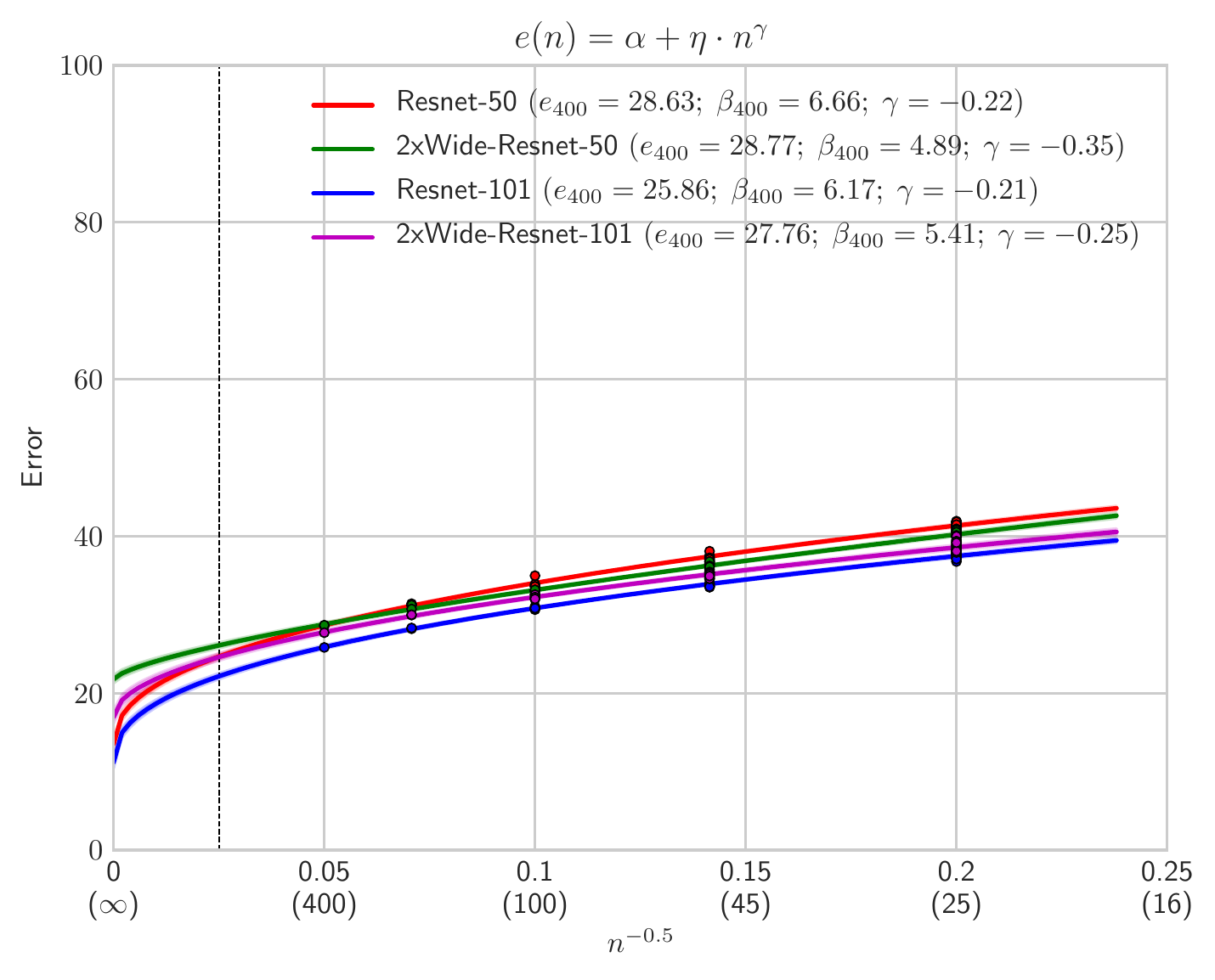}
        \caption{Width: linear}
        \label{fig:width_pretr_linear}
    \end{subfigure}
    % table
    \begin{subfigure}[t]{0.31\textwidth}
        \includegraphics[width=\textwidth]{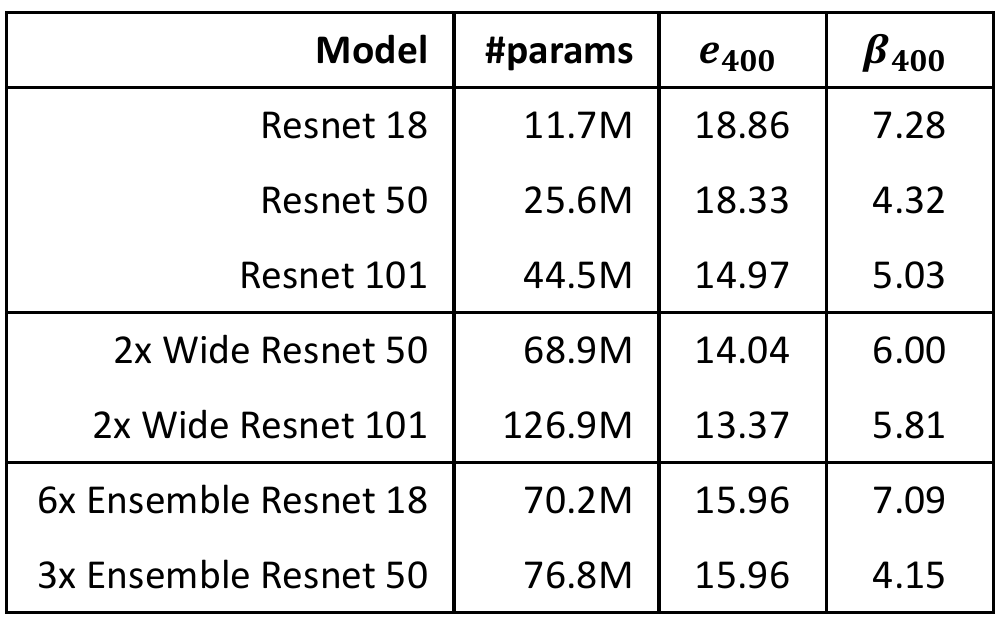}
        \caption{Ensemble vs Depth vs Width: finetune}
        \label{fig:ensemble_vs_depth_vs_width}
    \end{subfigure}
    \caption{
    \vspace{-0.05in}
    \label{fig:depth_width_ensemble}
    \textbf{Depth, width, and ensembles} on Cifar100.  The table (f) compactly compares several curves. %We tabulate error and data-reliance for seven settings in (f) to simplify comparison.
    } \vspace{-0.1in}
\end{figure*}
\textbf{Network depth, width, and ensembles: } The classical view is that smaller datasets need simpler models to avoid overfitting. In Figs.~\ref{fig:depth_pretr_ft},~\ref{fig:depth_pretr_linear}, we show that, not only do deeper networks have better potential at higher data sizes, their data reliance does not increase (nearly parallel and drops a little for fine-tuning), making deeper networks perfectly suitable for smaller datasets.  For linear classifiers (Fig.~\ref{fig:depth_pretr_linear}), the deeper networks provide better features, leading to consistent drop in $e_{400}$.  The small jump in data reliance between Resnet-34 and Resnet-50 may be due to the increased last layer input size from 512 to 2048 nodes.  When increasing width, the fine-tuned networks (Fig.~\ref{fig:width_pretr_ft}) have reduced $e_{400}$ without much change to data-reliance.  With linear classifiers (Fig.~\ref{fig:width_pretr_linear}), increasing the width leads to little change or even increase in $e_{400}$ with slight decrease in data-reliance.  

%. For example, \cite{rosenfeld2020icml_generalizationError} model generalization error as a sum of irreducible error and two power-law terms, one each for model size and data size, which enables computing an ideal model size given the data size but requires many more experiments to fit the curve parameters. 

An alternative to using a deeper or wider network is forming ensembles. Figure~\ref{fig:exp_ensemble} shows that, while an ensemble of six ResNet-18's (each 11.7M parameters) improves over a single model, it has higher $e_{400}$ and data-reliance than ResNet-101 (44.5M), Wide-ResNet-50 (68.9M), and Wide-ResNet-101 (126.9M). Three ResNet-50's (each 25.6M) underperforms Wide-ResNet-50 on $e_{400}$ but outperforms for small amounts of data due to lower data reliance. Fig.~\ref{fig:ensemble_vs_depth_vs_width} tabulates data reliance and error to simplify comparison.
%\zl{I realized comparing 6x18 vs wide-50 is unfair since the depths are different. Running Resnet-50 ensembles to compare to Wide-ResNet-50???}

\citet{rosenfeld2020icml_generalizationError} show that error can be modeled as a function of either training size, model size, or both.  Modeling both jointly can provide additional capabilities such as selecting model size based on data size, but requires many more experiments to fit the curve.  Our experiments show more clearly the effect on data reliance due to different ways of changing model size.

%\input{figs/effect_of_depth}

%\input{figs/effect_of_width}

%\input{figs/effect_of_ensemble} 

% Does a deeper network outperform an ensemble of shallower networks? 

% Train ensemble on the same data with variance coming from different initializations.

% \begin{figure*}[h!]
%     \centering
%     \begin{subfigure}[b]{0.32\textwidth}
%         \includegraphics[width=\textwidth]{imgs/aug_no_pretr_ft.pdf}
%         \caption{w/o Pretrain; Finetune}
%         \label{fig:aug_nopretr_ft}
%     \end{subfigure}
%     \begin{subfigure}[b]{0.32\textwidth}
%         \includegraphics[width=\textwidth]{imgs/aug_pretr_ft.pdf}
%         \caption{Pretrain; Finetune}
%         \label{fig:aug_pretr_ft}
%     \end{subfigure}
%     \begin{subfigure}[b]{0.32\textwidth}
%         \includegraphics[width=\textwidth]{imgs/aug_pretr_linear.pdf}
%         \caption{Pretrain; Linear}
%         \label{fig:aug_pretr_linear}
%     \end{subfigure}
% \caption{\textbf{Data augmentation} on Places365.}
% \label{fig:aug}
% \end{figure*}

\begin{figure}[h!]
    \centering
    \includegraphics[width=\columnwidth]{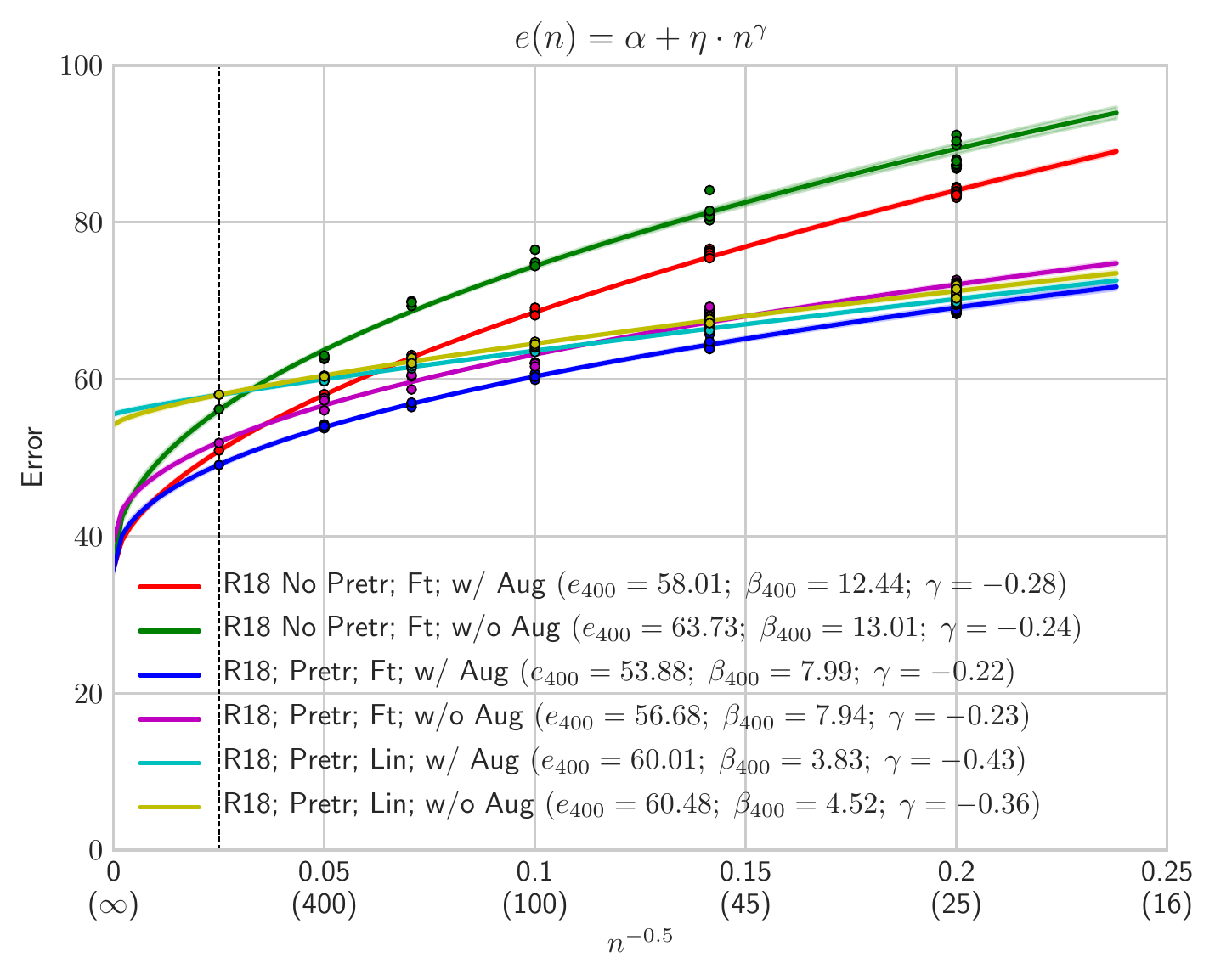}  
    \vspace{-0.3in}
\caption{\textbf{Data augmentation} on Places365.} \vspace{-0.2in}
\label{fig:aug}
\end{figure}
\textbf{Data Augmentation: } 
We are not aware of previous studies on interaction between data augmentation and training size. For example, a large survey~\cite{connor_augmentation_bigdata_2019} compares different augmentation methods only on full training size.  One may expect that data augmentation acts as a regularizer with reduced effect for large training sizes, or even possibly negative effect due to introducing bias. However, Fig.~\ref{fig:aug} shows that data augmentation on Places365 reduces error for all training sizes with little or no change to data-reliance when fine-tuning.  $e(n)$ with augmentation roughly equals $e(1.8n)$ without it, supporting the view that augmentation acts as a multiplier on the value of an example. For the linear classifier, data augmentation has little apparent effect due to low data-reliance, but the results are still consistent with this multiplier.

\section{Discussion}
\label{sec:discussion}

Evaluation methodology is the foundation of research, impacting how we choose problems and rank solutions. Large train and test sets now serve as the fuel and crucible to refine machine learning methods. %The current evaluation standard of using fixed i.i.d. train/test sets has supported many classification model improvements, but as machine learning broadens to continual learning, representation learning, long-tail learning, and so on, we need evaluation methods that better reflect the uncontrollable, unpredictable, and ever-changing world.  
Our experiments show that characterizing performance in terms of error and data-reliance yields a more complete understanding of model design and training size impact than single-point error. With that perspective, we discuss the limitations of our experiments and directions for future work.

%\begin{itemize}
    %\item 
    \textbf{Cause and impact of $\gamma$:} We speculate that $\gamma$ is largely determined by hyperparameters and optimization rather than model design. This presents an opportunity to identify poor training regimes and improve them.  Intuitively, one would expect that more negative $\gamma$ values are better (i.e. $\gamma=-1$ preferable to $\gamma=-0.5$), since the error is $O(n^\gamma)$, but we find the high-magnitude $\gamma$ tends to come with high asymptotic error, indicating that the efficiency comes at cost of over-commitment to initial conditions.  We speculate (but with some disagreement among authors) that $\gamma\approx-0.5$ is an indication of a well-trained curve and will generally outperform curves with higher or lower $\gamma$, given the same classification model. %It would be interesting to examine the impact of hyperparameter selection and optimization method on $\gamma$.
    
    %\item 
    \textbf{Small training sets}: Error is bounded and classifier performance with small training sets may be modeled as transitioning from random guess to informed prediction, as shown by \citet{rosenfeld2020icml_generalizationError}. For simplicity, we do not model performance with very small training size, but 
    %, partly to keep our model simple, partly because small training performance can be easily measured empirically, and partly because performance with small training size is highly variable depending on the sample. However, 
    studying the small data regime could be interesting, particularly to determine whether design decisions have an impact at the small size that is not apparent at larger sizes.
    
    %\item 
    \textbf{Losses and Prediction types}: We analyze multiclass classification error, but the same analysis could likely be extended to other prediction types.  For example, \citet{kaplan2020scaling} analyze learning manifolds of cross-entropy loss of language model transformers. Learning curve models of cross-entropy loss could also be used for problems like object detection or semantic segmentation that typically use more complex aggregate evaluations. 
    %may also be suitable for problems like object detection or grounding that have
    %Problems like object detection or grounding sometimes have relatively complex evaluation measures, such as average precision after accounting for localization and label accuracy, but test evaluation of the same losses used for training should still apply. \cite{sun_iccv2017_unreasonable_effectiveness} show an approximately log-linear behavior between mean intersection of union semantic segmentation error as a function of number of training samples.
    
    %\item 
    \textbf{More design parameters and interactions}: The interaction between data scale, model scale, and performance is well-explored by \citet{kaplan2020scaling} and \citet{rosenfeld2020icml_generalizationError}, but it could also be interesting to explore interactions, e.g. between class of architecture (e.g. VGG, ResNet, EfficientNet~\citep{tan2019efficientnet}) and some design parameters, to see the impact of ideas such as skip-connections, residual layers and bottlenecks.  More extensive evaluation of data augmentation, representation learning, optimization, and regularization would also be interesting.  
    %\item \textbf{Learning curve parameterization} We find that just two parameters is sufficient to model performance of a single classifier (defined by architecture, hyperparameters, and optimization method and parameters) with varying numbers of samples, if we use recent optimization techniques encapsulated by the Ranger library~\cite{ranger}. 
    
    %\item 
    \textbf{Unbalanced class distributions}: In most of our experiments, we use equal number of samples per class.  Further experimentation is required to determine whether class imbalance impacts the form of the learning curve.
%\end{itemize}

%\textbf{Limitations: } Our work in this paper is limited to classification loss, and our model does not account for small sample effects where chance performance is a major factor. Although our proposed $e_N$ and $\beta_N$ are stable under perturbations and different learning curve parameterizations, the asymptotic error $\alpha$ and exponent $\gamma$ parameters of the learning curve are unstable, and our confidence interval does not account for $\gamma$ variance.  Unstable $\alpha$ means that little can be concluded about asymptotic performance, though $e_N-\beta_N$ can stand in as a measure of large-data performance.  Unstable $\gamma$ may mean that conclusions are subject to the hyperparameter selection and optimization method.

%\textbf{Future work: } Do the hyperparameters such as learning rate, schedule, and weight decay determine $\gamma$, or something else? It appears that $\gamma<-0.5$ is accompanied by high $\alpha$ and/or $\eta$. Should $\gamma=-0.5$ for a well-trained system? Answering these questions could lead to improved training and evaluation methodologies.  It would also be interesting to investigate learning curve models for small training size, other losses and prediction types, more design parameters and interactions, and impact of imbalance in class distribution. 

%Appendix~\ref{app:discussion} offers extended discussion. 
The \textbf{supplemental material} contains implementation details (Appendix~\ref{supp:implementation}); a user guide to fitting, displaying, and using learning curves (Appendix~\ref{app:user_guide}); experiments on additional datasets (Appendix~\ref{app:extra}); and a table of learning curve parameters for all experiments, also comparing $e_N$ and $\beta_N$ produced by two learning curve models (Appendix~\ref{app:big_table}).
%provides a guide to fitting, displaying, and using learning curves.  contains a table of learning curves for all of our experiments and compares $e_N$ and $\beta_N$ produced by two learning curve models.

\section{Conclusion}

Performance depends strongly on training size, so size-variant analysis more fully shows the impact of innovations.  A reluctant researcher may protest that such analysis is too complicated, too computationally expensive, or requires too much space to display.  We show that learning curve models can be easily fit (Sec.~\ref{sec:estimation}) from a few trials (Fig.~\ref{fig:validity}) and compactly summarized (Fig.~\ref{fig:ensemble_vs_depth_vs_width}), leaving our evasive experimenter no excuse. Our experiments serve as examples that learning curve analysis can yield interesting observations. Learning curves can further inform training methodology, continual learning, and representation learning, among other problems, providing a better understanding of contributions and ultimately leading to faster progress in machine learning.

%classifier design decisions impact, not only performance at a single data size, but how quickly performance changes with more or fewer samples, and evaluation of optimizers, augmenters, etc. should t

%classifier performance depends strongly on training size, and the impact of newly proposed optimizers, augmenters, etc. mus
%It is clear that classifier performance depends heavily on training size, and to more 
%Classifier performance is best evaluated as a function of training size, but the literature lacks a standard methodology for such evaluation.  We show how to use learning curves to evaluate classifier design decisions. We propose to characterize the curve 
%We find an extended power law provides the best fit across many different architectures, datasets, and other design parameters.  
%with error $e_N$ and data-reliance $\beta_N$, which are stable under data perturbations and can be derived from different learning curve models. Our experiments lead to several interesting observations about impacts of pretraining, fine-tuning, data augmentation, depth, width, and ensembles. We anticipate learning curves can further inform training methodology, continual learning, and representation learning, among other problems, and hope to see learning curves become part of a standard classification evaluation.

%{\small
%\bibliographystyle{ieee_fullname}
%\bibliography{main,related_work}
%}

\bibliography{main,related_work}
\bibliographystyle{icml2021}
\storecounter{figure}{figurecounterstore}
\storecounter{table}{tablecounterstore}

\clearpage
\appendix
\pagenumbering{arabic} 

\twocolumn[
\icmltitle{ Learning Curves for Analysis of Deep Networks: Appendix}
]

\section{Implementation Details}
\label{supp:implementation}

We use Pytorch-Lightning~\citep{falcon2019pytorch} for our implementation with various architectures, weight initializations, data augmentation, and linear or fine-tuning optimization.

\textbf{Training: } We train models with images of size $224\times 224$ for all experiments (to facilitate use of pretrained models) with a batch size of 64 (except for Wide-ResNet101 and Wide-ResNeXt101, where we use a batch size of 32 and performed one optimizer step every two batches). For each experiment setting, we conduct a learning rate search on a subset of the training data and choose the learning rate with the highest validation accuracy, and use it for all other subsets. We determine each fold's training schedule on a mini-train/mini-val split of 2:1 on the train set. Each time the mini-val error stops decreasing for some epochs (``patience''), we revert to the best epoch and decrease the learning rate to 10\%, and we perform this twice. Then we use this optimal mini-train learning rate schedule and ending epoch to train on the whole fold. The patience is $\propto 1/\sqrt{n}$, and is 5 at the $n=400$ samples/class for CIFAR100/Places365 and 15 at the largest training size for other smaller datasets. % For Places365, Cifar100 and Cifar10, we used a patience value of 5 and for the remaining datasets we used a patience value of 15. 
We use a weight decay value of 0.0001. We use the Ranger optimizer~\citep{ranger}, 
%version 20.4.11, 
which combines Rectified Adam~\citep{liu2019variance}, Look Ahead~\citep{Zhang2019LookaheadOK}, and Gradient Centralization~\citep{Yong2020GradientCA}. In early experiments, we found Ranger to lead to lower error and to reduce sensitivity of hyperparameters, compared to vanilla SGD or Adam~\citep{kingma_iclr2015_adam}.

%\derek{Other hyperparameters?  E.g. weight decay?}

\textbf{Backbone Architecture: } We use the default Pytorch implementations of all of the following architectures: AlexNet~\citep{krizhevsky2012}, ResNet-18, ResNet-50, ResNet-101~\citep{he2015residual}, ResNeXt-50, ResNeXt-100~\citep{XieGDTH16}, VGG16\_BN~\citep{SimonyanZ14a}, Wide-ResNet-50, and Wide-ResNet-101~\citep{ZagoruykoK16}. For each architecture, we modify the last layer to match the same number of classes as the test dataset with Kaiming initialization~\citep{he2015delving}. 

\textbf{Number of Training Examples: } To compute learning curves for CIFAR and Places365, we vary the number of training examples per class, partition the train set, and train one model per partition.  For CIFAR100~\citep{cifar100}, we use $\{25, 50, 100, 200, 400\}$ training examples per class, and the number of models trained for each respectively is $\{16, 8, 4, 2, 1\}$. Similar to \cite{hestness_2017arxiv_deeplearningpredictable}, we find training sizes smaller than 25 samples per class are strongly influenced by bounded error and deviate from our model.  For Places365 dataset, 
we use $\{25, 50, 100, 200, 400, 1600\}$ training examples per class and $\{16, 8, 4, 3, 3, 1\}$ models each. For other datasets (Fig.~\ref{fig:additional_datasets}), we use $\{20\%, 40\%, 80\%\}$ of the full data and train $\{4, 2, 1\}$ models each.
% we use the same number of training examples per class as CIFAR, except that we also train one model with 1600 examples and train three models on separate partitions of sizes 200 and 400.  
We hold out 20\% of data from the original training set for testing (a validation set could also be used if available) to discourage meta-fitting on the test set.  For example, we hold out 100 samples per class from the original CIFAR100 training set and perform hyperparameter selection and training on the remaining 400 samples.  
% In general, we hold out 20\% data from the original validation set (if available) or training set (if not) for testing to discourage meta-fitting on the test set.  For example, for CIFAR, which has no validation set, our test set is 100 held out samples per class from the original training set, and training and hyperparameter selection is done on the remaining 400 samples.

\textbf{Pretraining: } When pretraining is used, we initialize models with pretrained weights learned through supervised training on ImageNet or Places365, or MOCO self-supervised training on ImageNet~\citep{he2020moco}. Otherwise, weights are randomly initialized with Kaiming initialization. 

\textbf{Data Augmentation: } For CIFAR, we pad by 4 pixels and use a random $32\times 32$ crop (test without augmentation), and for Places365 we use random-sized crop~\citep{szegedy2015going} to 224$\times$224 and random flipping (center crop 224$\times$224 test time). For remaining datasets, we follow the pre-processing in \citet{zhai2020largescale} that produced the best results when training from scratch.

\textbf{Linear vs. Fine-tuning: } For ``linear'', we only train the final classification layer, with the other weights frozen to initialized values.  All weights are trained when ``fine-tuning''.

\section{User's Guide to Learning Curves}
\label{app:user_guide}

\subsection{Uses for Learning Curves}

\begin{itemize}
    \item \textbf{Comparison}: When comparing two learners, measuring the error and data-reliance provides a better understanding of the differences than evaluating single-point error. We compare curves with $e_N$ and $\beta_N$, rather than directly using the curve parameters, because they are more stable under data perturbations and do not depend on the parameterization, instead corresponding to error and rate of change about $n=N$.  The difference $e_N-\beta_N$ can be used as a measure of large-sample performance.
    \item \textbf{Performance extrapolation}: A 10x increase in training data can require a large investment, sometimes millions of dollars. Learning curves can predict how much performance will improve with the additional data to judge whether the investment is worthwhile. 
    \item \textbf{Model selection}: When much training data is available, architecture, hyperparameters, and losses can be designed and selected using a small subset of the data to minimize the extrapolated error of the full training set size. Higher-parameter models such as in \cite{kaplan2020scaling} and \cite{rosenfeld2020icml_generalizationError} may be more useful as a mechanism to simultaneously select scale parameters and extrapolate performance, though fitting those models is much more computationally expensive due to the requirement of sampling error/loss at multiple scales and data sizes.
    \item \textbf{Hyperparameter validation}: A poor fitting learning curve (or one with $\gamma$ far from $-0.5$) is an indication of poor choice of hyperparameters, as pointed out by \cite{hestness_2017arxiv_deeplearningpredictable}.
\end{itemize}

\subsection{Estimating and Displaying Learning Curves}

\textbf{Use validation set: } We recommend computing learning curves on a validation set, rather than a test set, according to best practice of performing a single evaluation on the test set for the final version of the algorithm. All of our experiments are on a validation set, which is carved from the official training set if necessary.

\begin{figure}[h!]
\centering
\includegraphics[width=\linewidth]{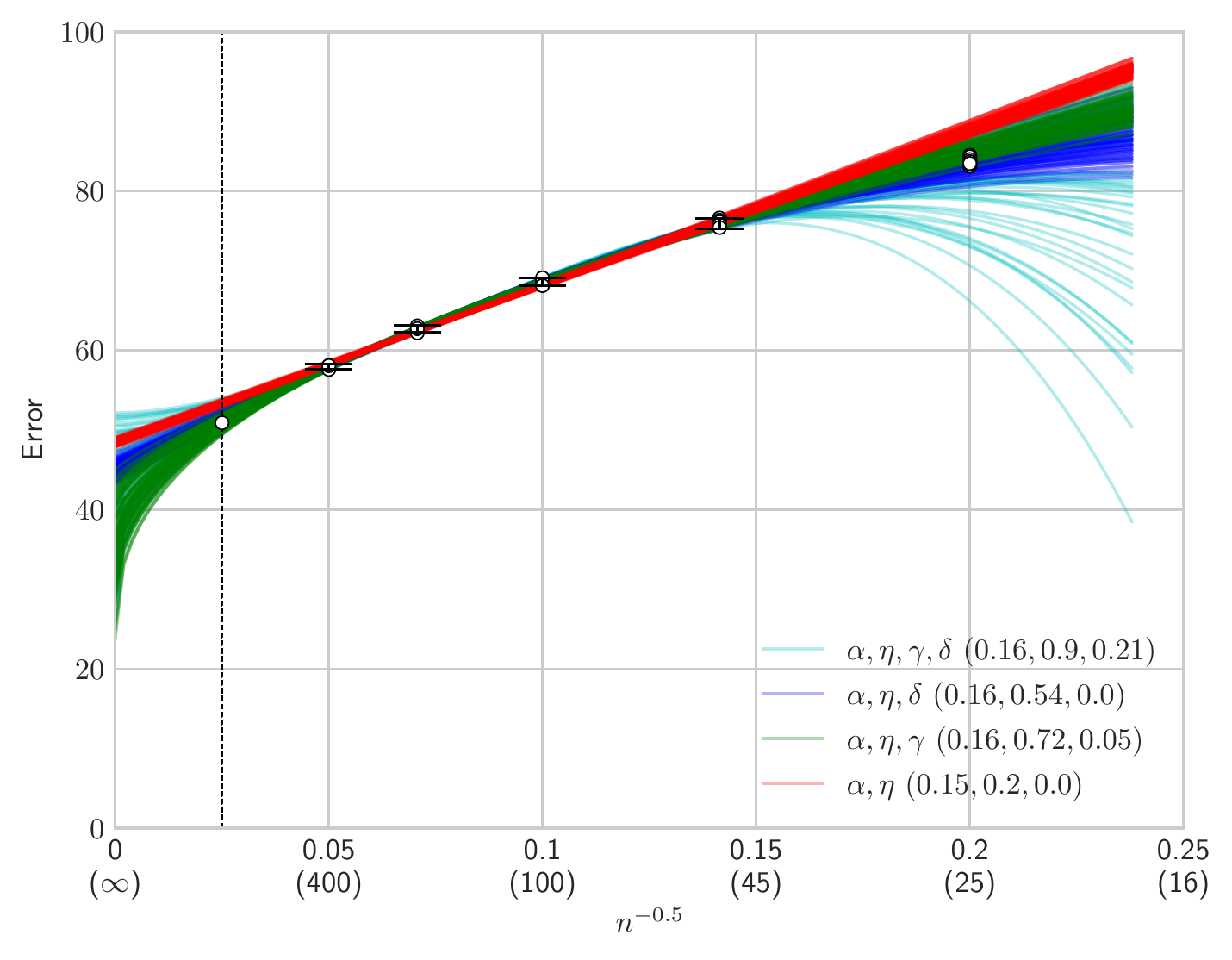}
\caption{
\label{fig:stability_measurements}
Stability under sparse measurements: Sampled learning curves for Places365 fine-tuned without pretraining are shown for four different learning curve parameterizations.  In each case, means and standard deviations (shown by error bars) are estimated for $n=50$, $n=100$, $n=200$, $n=400$, using all the data points shown as white circles.  Then, 100 times, we sample one point each from a Guassian distribution and fit a learning curve to the four points.  In parantheses, the legend shows the standard deviation of $e_N$, $\beta_N$, and $\gamma$. Note that the parameterization of $\{\alpha, \eta, \gamma\}$ extrapolates best to lower and higher data sizes while still producing stable estimates of $e_N$ and $\beta_N$.  Asymptotic error, however, varies widely.
}
\end{figure}

\textbf{Generate at least four data points: } In most of our experiments on CIFAR100, we train a 31 models: 1 on 400 images, 2 on 200 images, 4 on 100 images, 8 on 50 images, and 16 on 25 images.  Each trained model provides one data point, the average validation error.  In each case, the training data is partitioned so that the image sets within the same size are non-overlapping.  Training multiple models at each size enables estimating the standard deviation for performing weighted least squares and producing confidence bounds.  However, our experiments indicate that learning curves are highly stable, so a minimal experiment of training four models on the full, half, quarter, and eighth-size training set may be sufficient as part of a standard evaluation.  See Fig.~\ref{fig:stability_measurements}  It may be necessary to train more models if attempting to distinguish fine differences.

\textbf{Set hyperparameters: } The learning rate and learning schedule are key parameters to be set. We have not experimented with changes to weight decay, momentum, or other hyperparameters.

\textbf{Fit learning curves: } If more than one data point is available for the same training size, the standard deviation can be estimated. As described in Sec.~\ref{sec:estimation}, we recommend fitting a model of $\sigma_i^2 = \sigma_0^2 + \hat\sigma^2/n$, where $\sigma_0^2$.  $\sigma_0^2$ is the variance due to randomness in initialization and optimization.  The fitting is not highly sensitive to this parameter, so we recommend setting $\sigma_0^2=0.01$ and fitting $\hat\sigma$ to observations, since estimating both from experiments to generate a single learning curve introduces high variance and instability.

\textbf{Display learning curves or parameters: } As in this paper, learning curves can be plotted linearly with the x-axis as $n^{-0.5}$ and the y-axis as error. We choose this rather than log-linear because it reveals prediction of asymptotic error and yields a linear plot when $\gamma=-0.5$.
Since space is often a premium, the learning curve parameters can be displayed instead, as illustrated in Table~\ref{tab:lc_table}. Although $\gamma$ is not useful for direct comparison, including it enables recovery of $\alpha$, $\eta$, and $\gamma$ to plot the original learning curve.

\begin{table}[h]
\caption{\label{tab:lc_table}
Results: $\rm model_1$ and $\rm model_2$ have similar percent test error when training on the full set.  Fitting a learning curve on the validation set, we see that $\rm model_2$ has higher data-reliance, so may outperform for larger training sets. {\em This is a hypothetical example to illustrate use of learning curves in a table.}
}
  \centering
  \begin{tabular}{lccc}
    \toprule
    & $e_N$ & $\beta_N$ & $\gamma$ \\
    \cmidrule{2-4}
    $\rm model_1$ & 25.3 \% & 4.6 & -0.36 \\
    $\rm model_2$ & 25.2 \% & 8.4 & -0.47 \\
    \bottomrule
    \end{tabular}
\end{table}

\section{Additional Results}
\label{app:extra}

\begin{figure}[h!]
    \centering
    \includegraphics[width=\linewidth]{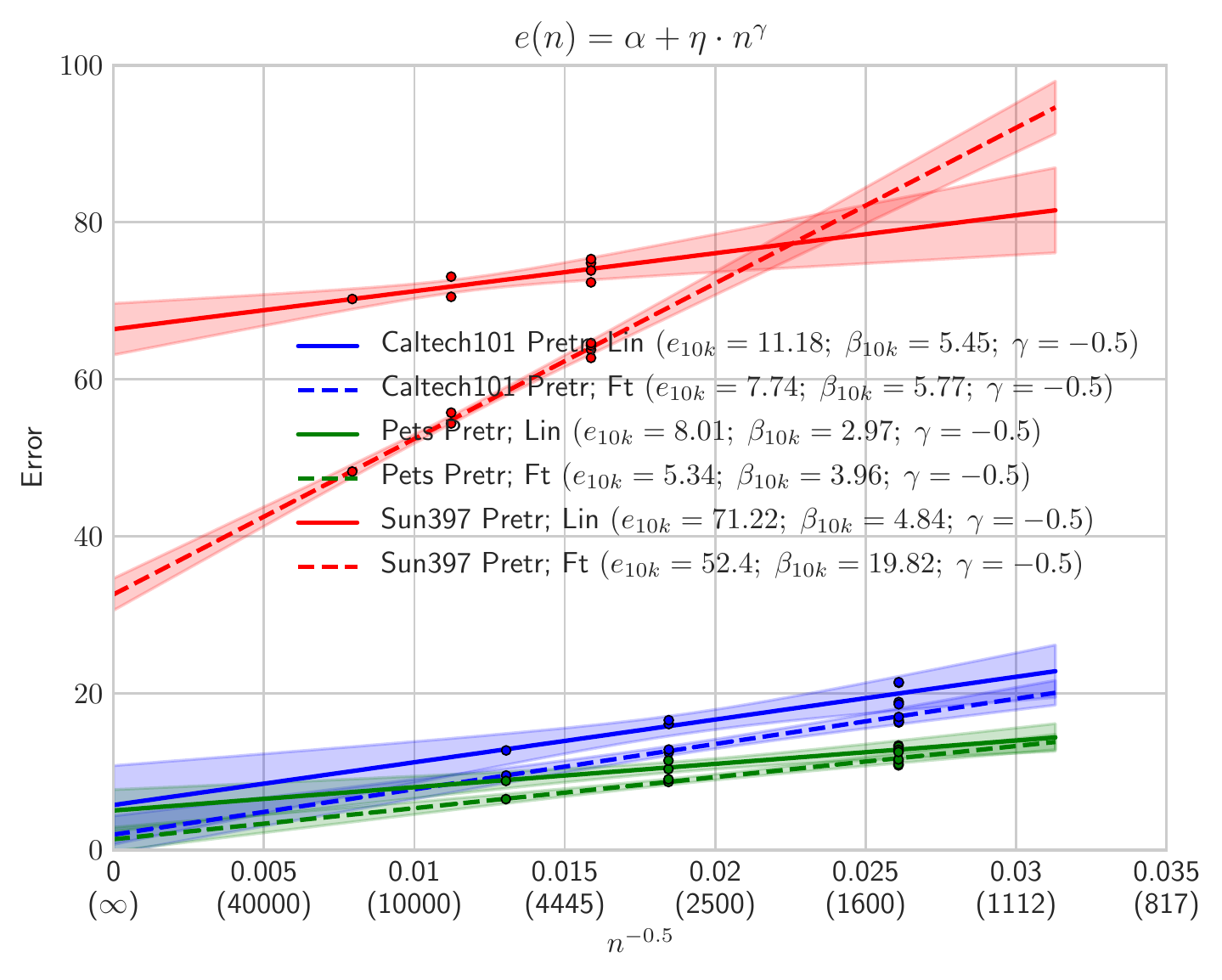}
    \vspace{-0.3in}
    \caption{
    \label{fig:additional_datasets}
    \textbf{Additional datasets}
    \vspace{-0.1in}
    }
\end{figure}

% \begin{figure}[ht]
%     \captionsetup[subfigure]{font=footnotesize,labelfont=scriptsize}
%     \centering
%     % dataset
%     \begin{subfigure}[t]{0.48\textwidth}
%         \includegraphics[width=\textwidth]{imgs/dataset_ft.pdf}
%         \caption{Datasets: effect of finetuning}
%         \label{fig:datasets_ft}
%     \end{subfigure}
%     \begin{subfigure}[t]{0.48\textwidth}
%         \includegraphics[width=\textwidth]{imgs/dataset_pretr.pdf}
%         \caption{Datasets: effect of pretraining}
%         \label{fig:datasets_pr}
%     \end{subfigure}

%     \caption{
%     \label{fig:additional_datasets}
%     \textbf{Additional datasets} 
%     %\textbf{Learning Curves:} Each plot shows the percent error as the y-axis and $1/\sqrt(n)$ as the x-axis, with $n$ values in parentheses showing the number of samples per class (unless noted otherwise). The y-intercept is asymptotic error at $n=\infty$. %The curve is unbounded at the right side, but we use a minimum of 25 training samples per class because small training sample sizes lead to high variance on test error and inaccuracies due to upper-bounded chance error. 
%     %Unless noted, all tests are on Cifar100 with pretraining on ImageNet. Comparison on linear (training only classification layer) highlights the quality of features learned in pretraining.
%     }
% \end{figure}
\textbf{Additional datasets: }  In Fig.~\ref{fig:additional_datasets}, we verify that our learning curve model fits to multiple other datasets (chosen from natural tasks in \cite{zhai2020largescale}), comparing fine-tuned vs. linear with Resnet-18.  For these plots only, $n$ is the total number of samples. The $\gamma$ values are estimated from data, but the prior has more effect here due to fewer error measurements.
We see fine-tuning consistently outperforms linear, though the difference is most dramatic for Sun397. %Pretraining provides large benefits across datasets.  

%\subsection{Full size figures}
%Due to space restrictions and to simplify comparisons between plots, we include small versions of the figures in the main section.  Here, we include larger figures which may be easier to view in printouts.
\section{Table of Learning Curves}
\label{app:big_table}
Table~\ref{tab:bigtable} shows experimental settings and fit parameters for learning curves under two parameterizations.  We can see that similar $e_{400}$ and $\beta_{400}$ values are obtained when fixing $\gamma=-0.5$ and fitting to errors with only three training sizes (RMS difference in $e_{400}$ and $\gamma_{400}$ are 0.42 and 0.95, respectively).  This means that learning curves can be fit and compared without training a large number of additional models.

\begin{table*}[h!]
\caption{
\label{tab:bigtable}
Experiment settings and parameters: We show the datasets, architectures, settings, and learning rate (set by mini-train/val) used to train and test our classifiers.  Next, we show the parameters fit using the extended power law model $e(n)=\alpha+\eta n^{-\gamma}$.  Next to that, we show the model resulting from setting $\gamma=-0.5$ and fitting to only the three training sizes with highest n. }
\begin{center}
\resizebox{\linewidth}{!}{%
\begin{tabular}{lllllccl|rrrrr|rrrr}
\toprule
\multicolumn{1}{l}{\textbf{}} & \multicolumn{1}{l}{\textbf{}} & \multicolumn{1}{l}{\textbf{}} & \multicolumn{1}{l}{\textbf{}} & \multicolumn{1}{l}{\textbf{}}     & \multicolumn{1}{l}{\textbf{}}  & \multicolumn{1}{l}{\textbf{}} & \multicolumn{1}{l}{\textbf{}} & \multicolumn{5}{c}{\textbf{extended power law}} & \multicolumn{4}{c}{$n^{-0.5}$ \textbf{linear fit to last 3 points}} \\
\multicolumn{1}{l}{}          & \multicolumn{1}{l}{\textbf{dataset}}   & \multicolumn{1}{l}{\textbf{arch}}      & \multicolumn{1}{l}{\textbf{\# param}}  & \multicolumn{1}{l}{\textbf{pretrain/init}} & \multicolumn{1}{l}{\textbf{fine-tune?}} & \multicolumn{1}{l}{\textbf{data aug?}} & \multicolumn{1}{l}{\textbf{lrnRate}}   & $\alpha$    & $\eta$   & $\gamma$   & $e_{400}$   & $\beta_{400}$   & $\alpha$              & $\eta$              & $e_{400}$              & $\beta_{400}$              \\
\midrule
\multicolumn{17}{l}{PRETRAIN\_IN2CIFAR}                                                                                                                                                                                                                                                                                                                                                                        \\
No Pretr; Linear            & CIFAR     & Resnet-18           & 51K       & Random          & No         & Yes       & 0.01                          & 78.51                     & 120.13                  & -0.84                     & 79.29                    & 1.32                        & 78.06                     & 26.12                   & 79.36                    & 1.31                        \\
No Pretr; Finetune          & CIFAR     & Resnet-18           & 11.7M     & Random          & Yes        & Yes       & 0.01                          & 5.68                      & 259.29                  & -0.41                     & 27.91                    & 18.23                       & 11.21                     & 336.13                  & 28.02                    & 16.81                       \\
Pretr; Linear               & CIFAR     & Resnet-18           & 51K       & ImageNet        & No         & Yes       & 0.0003                        & 24.4                      & 65.28                   & -0.35                     & 32.42                    & 5.61                        & 27.33                     & 102.16                  & 32.44                    & 5.11                        \\
Pretr; Finetune             & CIFAR     & Resnet-18           & 11.7M     & ImageNet        & Yes        & Yes       & 0.001                         & 12.48                     & 194.19                  & -0.57                     & 18.86                    & 7.28                        & 11.37                     & 150.73                  & 18.91                    & 7.54                        \\ \midrule
\multicolumn{17}{l}{PRETRAIN\_IN2PLACES}                                                                                                                                                                                                                                                                                                                                                                       \\
No Pretr; Linear            & Places    & Resnet-18           & 187K      & Random          & No         & Yes       & 0.03                          & 91.84                     & 19.13                   & -0.5                      & 92.79                    & 0.96                        & 91.09                     & 29.31                   & 92.55                    & 1.47                        \\
No Pretr; Finetune          & Places    & Resnet-18           & 11.7M     & Random          & Yes        & Yes       & 0.001                         & 33.16                     & 117.45                  & -0.26                     & 57.89                    & 12.86                       & 44.39                     & 263.63                  & 57.57                    & 13.18                       \\
Pretr; Linear               & Places    & Resnet-18           & 187K      & ImageNet        & No         & Yes       & 0.0003                        & 54.43                     & 53.39                   & -0.38                     & 59.91                    & 4.16                        & 56.11                     & 76.95                   & 59.95                    & 3.85                        \\
Pretr; Finetune             & Places    & Resnet-18           & 11.7M     & ImageNet        & Yes        & Yes       & 0.0003                        & 40.92                     & 70.04                   & -0.28                     & 54                       & 7.33                        & 44.82                     & 174.69                  & 53.55                    & 8.73                        \\ \midrule
\multicolumn{17}{l}{PRETRAIN\_IN\_PLACES\_MOCO2CIFAR}                                                                                                                                                                                                                                                                                                                                                          \\
No Pretr                    & CIFAR     & Resnet-50           & 25.6M     & Random          & Yes        & Yes       & 0.01                          & -2.23                     & 243.44                  & -0.35                     & 27.66                    & 20.93                       & 9.18                      & 372.11                  & 27.79                    & 18.61                       \\
Pretr on Imagenet           & CIFAR     & Resnet-50           & 25.6M     & ImageNet        & Yes        & Yes       & 0.001                         & 15.11                     & 178.61                  & -0.67                     & 18.33                    & 4.32                        & 13.3                      & 99.88                   & 18.29                    & 4.99                        \\
Pretr on Places             & CIFAR     & Resnet-50           & 25.6M     & Places          & Yes        & Yes       & 0.001                         & -5.61                     & 109.92                  & -0.24                     & 20.49                    & 12.53                       & 9.05                      & 195.43                  & 18.82                    & 9.77                        \\
Pretr on Imagenet with MOCO & CIFAR     & Resnet-50           & 25.6M     & ImageNet (MOCO) & Yes        & Yes       & 0.0003                        & 0.07                      & 112.69                  & -0.3                      & 18.74                    & 11.21                       & 10.07                     & 210.67                  & 20.61                    & 10.53                       \\ \midrule
\multicolumn{17}{l}{DEPTH\_FT}                                                                                                                                                                                                                                                                                                                                                                                 \\
Resnet-18                   & CIFAR     & Resnet-18           & 11.7M     & ImageNet        & Yes        & Yes       & 0.001                         & 12.48                     & 194.19                  & -0.57                     & 18.86                    & 7.28                        & 11.37                     & 150.73                  & 18.91                    & 7.54                        \\
Resnet-34                   & CIFAR     & Resnet-34           & 21.8M     & ImageNet        & Yes        & Yes       & 0.001                         & 15.76                     & 237.19                  & -0.73                     & 18.75                    & 4.36                        & 13.51                     & 104.15                  & 18.72                    & 5.21                        \\
Resnet-50                   & CIFAR     & Resnet-50           & 25.6M     & ImageNet        & Yes        & Yes       & 0.001                         & 15.11                     & 178.61                  & -0.67                     & 18.33                    & 4.32                        & 13.3                      & 99.88                   & 18.29                    & 4.99                        \\
Resnet-101                  & CIFAR     & Resnet-101          & 44.5M     & ImageNet        & Yes        & Yes       & 0.0003                        & 10.91                     & 166.44                  & -0.62                     & 14.97                    & 5.03                        & 8.95                      & 117.22                  & 14.81                    & 5.86                        \\ \midrule
\multicolumn{17}{l}{DEPTH\_LINEAR}                                                                                                                                                                                                                                                                                                                                                                             \\
Resnet-18                   & CIFAR     & Resnet-18           & 51K       & ImageNet        & No         & Yes       & 0.001                         & 24.4                      & 65.28                   & -0.35                     & 32.42                    & 5.61                        & 27.33                     & 102.16                  & 32.44                    & 5.11                        \\
Resnet-34                   & CIFAR     & Resnet-34           & 51K       & ImageNet        & No         & Yes       & 0.001                         & 21.77                     & 59.56                   & -0.33                     & 30.02                    & 5.44                        & 25.11                     & 98.33                   & 30.03                    & 4.92                        \\
Resnet-50                   & CIFAR     & Resnet-50           & 205K      & ImageNet        & No         & Yes       & 0.0003                        & 13.5                      & 56.54                   & -0.22                     & 28.63                    & 6.66                        & 23.05                     & 112.08                  & 28.65                    & 5.6                         \\
Resnet-101                  & CIFAR     & Resnet-101          & 205K      & ImageNet        & No         & Yes       & 0.0003                        & 11.17                     & 51.7                    & -0.21                     & 25.86                    & 6.17                        & 20.98                     & 99.32                   & 25.95                    & 4.97                        \\ \midrule
\multicolumn{17}{l}{WIDTH\_FT}                                                                                                                                                                                                                                                                                                                                                                                 \\
Resnet-50                   & CIFAR     & Resnet-50           & 25.6M     & ImageNet        & Yes        & Yes       & 0.001                         & 15.11                     & 178.61                  & -0.67                     & 18.33                    & 4.32                        & 13.3                      & 99.88                   & 18.29                    & 4.99                        \\
2xWide-Resnet-50            & CIFAR     & Wide\_Resnet-50\_2  & 68.9M     & ImageNet        & Yes        & Yes       & 0.0003                        & 8.78                      & 160.14                  & -0.57                     & 14.04                    & 6                           & 7.83                      & 124.55                  & 14.06                    & 6.23                        \\
Resnet-101                  & CIFAR     & Resnet-101          & 44.5M     & ImageNet        & Yes        & Yes       & 0.0003                        & 10.91                     & 166.44                  & -0.62                     & 14.97                    & 5.03                        & 8.95                      & 117.22                  & 14.81                    & 5.86                        \\
2xWide-Resnet-101           & CIFAR     & Wide\_Resnet-101\_2 & 126.9M    & ImageNet        & Yes        & Yes       & 0.0003                        & 7.56                      & 116.21                  & -0.5                      & 13.37                    & 5.81                        & 7.55                      & 116.26                  & 13.36                    & 5.81                        \\ \midrule
\multicolumn{17}{l}{WIDTH\_LINEAR}                                                                                                                                                                                                                                                                                                                                                                             \\
Resnet-50                   & CIFAR     & Resnet-50           & 205K      & ImageNet        & No         & Yes       & 0.0003                        & 13.5                      & 56.54                   & -0.22                     & 28.63                    & 6.66                        & 23.05                     & 112.08                  & 28.65                    & 5.6                         \\
2xWide-Resnet-50            & CIFAR     & Wide\_Resnet-50\_2  & 205K      & ImageNet        & No         & Yes       & 0.0001                        & 21.78                     & 56.88                   & -0.35                     & 28.77                    & 4.89                        & 24.45                     & 87.19                   & 28.81                    & 4.36                        \\
Resnet-101                  & CIFAR     & Resnet-101          & 205K      & ImageNet        & No         & Yes       & 0.0003                        & 11.17                     & 51.7                    & -0.21                     & 25.86                    & 6.17                        & 20.98                     & 99.32                   & 25.95                    & 4.97                        \\
2xWide-Resnet-101           & CIFAR     & Wide\_Resnet-101\_2 & 205K      & ImageNet        & No         & Yes       & 0.0003                        & 16.95                     & 48.35                   & -0.25                     & 27.76                    & 5.41                        & 23.27                     & 90.85                   & 27.81                    & 4.54                        \\ \midrule
\multicolumn{17}{l}{AUG\_NO\_PRETR\_FT}                                                                                                                                                                                                                                                                                                                                                                        \\
Resnet-18 w/ Data-Aug       & Places    & Resnet-18           & 11.7M     & Random          & Yes        & Yes       & 0.001                         & 35.79                     & 118.94                  & -0.28                     & 58.01                    & 12.44                       & 44.39                     & 263.63                  & 57.57                    & 13.18                       \\
Resnet-18 w/o Data-Aug      & Places    & Resnet-18           & 11.7M     & Random          & Yes        & No        & 0.003                         & 36.62                     & 114.2                   & -0.24                     & 63.73                    & 13.01                       & 48.84                     & 288.46                  & 63.26                    & 14.42                       \\ \midrule
\multicolumn{17}{l}{AUG\_PRETR\_FT}                                                                                                                                                                                                                                                                                                                                                                            \\
Resnet-18 w/ Data-Aug       & Places    & Resnet-18           & 11.7M     & ImageNet        & Yes        & Yes       & 0.0003                        & 35.72                     & 67.85                   & -0.22                     & 53.88                    & 7.99                        & 44.82                     & 174.69                  & 53.55                    & 8.73                        \\
Resnet-18 w/o Data-Aug      & Places    & Resnet-18           & 11.7M     & ImageNet        & Yes        & No        & 0.001                         & 39.43                     & 68.44                   & -0.23                     & 56.68                    & 7.94                        & 47.34                     & 183.99                  & 56.53                    & 9.2                         \\ \midrule
\multicolumn{17}{l}{AUG\_PRETR\_LINEAR}                                                                                                                                                                                                                                                                                                                                                                        \\
Resnet-18 w/ Data-Aug       & Places    & Resnet-18           & 187K      & ImageNet        & No         & Yes       & 0.0003                        & 55.55                     & 58.56                   & -0.43                     & 60.01                    & 3.83                        & 56.11                     & 76.95                   & 59.95                    & 3.85                        \\
Resnet-18 w/o Data-Aug      & Places    & Resnet-18           & 187K      & ImageNet        & No         & No        & 0.001                         & 54.2                      & 54.27                   & -0.36                     & 60.48                    & 4.52                        & 55.62                     & 96.08                   & 60.42                    & 4.8                         \\ \midrule
\multicolumn{17}{l}{ARCHITECTURES\_IN2CIFAR}                                                                                                                                                                                                                                                                                                                                                                   \\
AlexNet                     & CIFAR     & AlexNet             & 61.1M     & ImageNet        & Yes        & Yes       & 0.001                         & 7.52                      & 131.77                  & -0.32                     & 26.89                    & 12.4                        & 15.97                     & 219.36                  & 26.94                    & 10.97                       \\
VGG-16(bn)                  & CIFAR     & VGG-16BN            & 138.4M    & ImageNet        & Yes        & Yes       & 0.0003                        & 6.21                      & 125.57                  & -0.38                     & 19.1                     & 9.79                        & 10.07                     & 181.1                   & 19.13                    & 9.06                        \\
ResNet-50                   & CIFAR     & Resnet-50           & 25.6M     & ImageNet        & Yes        & Yes       & 0.001                         & 15.11                     & 178.61                  & -0.67                     & 18.33                    & 4.32                        & 13.3                      & 99.88                   & 18.29                    & 4.99                        \\
ResNeXt-50(32x4d)           & CIFAR     & ResNeXt-50(32x4d)   & 25.0M     & ImageNet        & Yes        & Yes       & 0.001                         & 12.67                     & 185.51                  & -0.65                     & 16.45                    & 4.91                        & 11.37                     & 102.91                  & 16.52                    & 5.15                        \\
ResNet-101                  & CIFAR     & Resnet-101          & 44.5M     & ImageNet        & Yes        & Yes       & 0.0003                        & 10.91                     & 166.44                  & -0.62                     & 14.97                    & 5.03                        & 8.95                      & 117.22                  & 14.81                    & 5.86                        \\ \midrule
\multicolumn{17}{l}{ENSEMBLE}                                                                                                                                                                                                                                                                                                                                                                                  \\
1xResnet-18                 & CIFAR     & Resnet-18           & 11.7M     & ImageNet        & Yes        & Yes       & 0.001                         & 12.48                     & 194.19                  & -0.57                     & 18.86                    & 7.28                        & 11.37                     & 150.73                  & 18.91                    & 7.54                        \\
6xResnet-18                 & CIFAR     & Resnet-18           & 70.1M     & ImageNet        & Yes        & Yes       & 0.001                         & 8.73                      & 136.26                  & -0.49                     & 15.96                    & 7.09                        & 8.55                      & 147.16                  & 15.91                    & 7.36                       \\          
1xResnet-50                 & CIFAR     & Resnet-50           & 25.6M & ImageNet        & Yes        & Yes       & 0.001                         & 15.11                     & 178.61                  & -0.67                     & 18.33                    & 4.32                        & 13.3                      & 99.88                   & 18.29                    & 4.99                        \\
3xResnet-50                 & CIFAR     & Resnet-50           & 76.7M & ImageNet        & Yes        & Yes       & 0.001                         & 12.72                     & 150.13                  & -0.64                     & 15.96                    & 4.15                        & 11.43                     & 90.59                   & 15.96                    & 4.53                       \\
\bottomrule
\end{tabular}
}
\end{center}
\end{table*}

\end{document}